\pdfoutput=1
\documentclass[11pt]{article}

\usepackage{acl}
\usepackage{times}
\usepackage{latexsym}
\usepackage{multirow}
\usepackage{comment}
\usepackage{paralist}
\usepackage{enumitem}
\usepackage{etaremune}
\usepackage{color,framed}
\usepackage{xcolor,colortbl}
\usepackage{caption}
\usepackage{subcaption}
\usepackage[T1]{fontenc}
\usepackage[utf8]{inputenc}
\usepackage{microtype}

\usepackage{graphicx}
\graphicspath{{graphics/}}
\usepackage{xspace}
\usepackage{booktabs}
\usepackage{arydshln}
\usepackage{amssymb}
\usepackage{amsmath}    % for {align}
\definecolor{shadecolor}{rgb}{0.92,0.92,0.92}
\definecolor{Gray}{gray}{0.95}

\usepackage[prependcaption,textsize=tiny]{todonotes}

\newcommand{\datasetname}{\textsc{TVShowGuess }}
\newcommand{\datasetnamens}{\textsc{TVShowGuess}}

\newcommand{\goutou}{%
  \begingroup\normalfont
  \includegraphics[height=\fontcharht\font`\B]{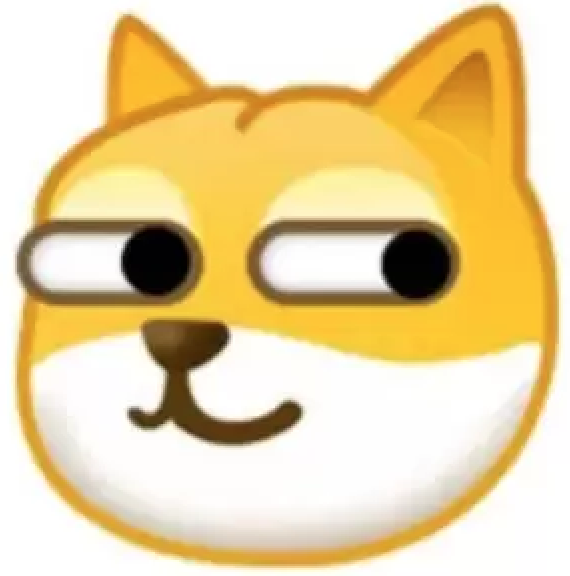}%
  \endgroup
}

\definecolor{clabel}{HTML}{929292}

\definecolor{cinput}{HTML}{6EB4FD}
\definecolor{coutput}{HTML}{F89151}
\definecolor{cgroundtruth}{HTML}{99C893}
\definecolor{cpink}{HTML}{FCD8D4}

\title{\datasetnamens: Character Comprehension in Stories\\ as Speaker Guessing}
  
\author{%
  Yisi Sang$^{1}$\thanks{\,\,Authors contributed equally to this paper. Mo Yu is the corresponding author.} \quad Xiangyang Mou$^{2*}$ \quad Mo Yu$^{3*}$ \quad Shunyu Yao$^{4}$ \quad Jing Li$^{5}$ \quad  Jeffrey Stanton$^{1}$ \\
  \small{$^1$Syracuse University \qquad $^2$Rensselaer Polytechnic Institute \qquad $^3$Pattern Recognition Center, WeChat AI} \\ \small{$^4$Princeton University \qquad $^5$New Jersey Institute of Technology} \\ 
  \small{\texttt{yisang@syr.edu}\quad \texttt{moux4@rpi.edu} \quad \texttt{moyumyu@tencent.com}}}

\begin{document}
\maketitle
\begin{abstract}
We propose a new task for assessing machines' skills of understanding fictional characters in narrative stories. The task, \datasetnamens, builds on the scripts of TV series and takes the form of guessing the anonymous main characters based on the backgrounds of the scenes and the dialogues. Our human study supports that this form of task covers comprehension of multiple types of character persona, including understanding characters' personalities, facts and memories of personal experience, which are well aligned with the psychological and literary theories about the theory of mind (ToM) of human beings on understanding fictional characters during reading. We further propose new model architectures to support the contextualized encoding of long scene texts. Experiments show that our proposed approaches significantly outperform baselines, yet still largely lag behind the (nearly perfect) human performance.
Our work serves as a first step toward the goal of narrative character comprehension.\footnote{Our code and data are released at \url{https://github.com/YisiSang/TVSHOWGUESS}.}

\end{abstract}

%%%%%%%%%%%%%%%%%%%%%%%%%%%%%%%%%%%%%%%%%%%%%%  
%%%%%%%%%%%%%%%  Introduction  %%%%%%%%%%%%%%%
%%%%%%%%%%%%%%%%%%%%%%%%%%%%%%%%%%%%%%%%%%%%%%

\section{Introduction}\label{sec:introduction}

Stories have two essential elements, plots and characters~\cite{mckee1997story}.
Character comprehension has been widely recognized as key to understanding stories in psychological, literary and educational research~\cite{bower1990mental,kennedy2013literature,currie2009narrative,paris2003assessing}.
\begin{figure}[t!]
    \fontsize{14}{10}\selectfont
    \centering
    \includegraphics[width=0.44\textwidth]{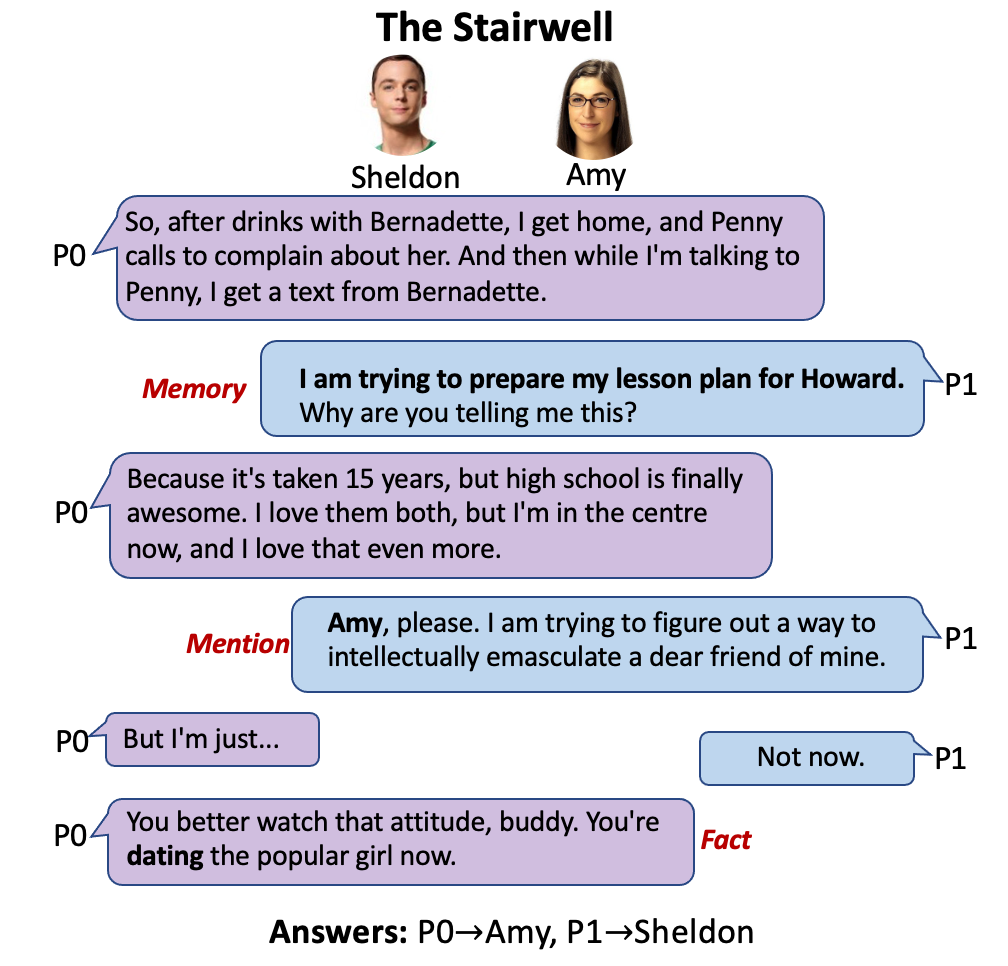}
    \vspace{-3mm}
    \caption{\small{A scene example from \datasetnamens. The character \emph{Amy} can be determined within the scene or with the fact of her relationship; while guessing \emph{Sheldon} would require memory of the character from previous episodes.}}
    \label{fig:bookqa_example}
    \vspace{-6mm}
\end{figure}
When reading stories, humans build mental models for characters based on their personae, which helps people to explain a character's emotional status~\cite{gernsbacher1998automatically}, identity, understand future behaviors~\cite{mead1990representation}, and even make counterfactual inferences for stories about that character~\cite{fiske1979imaging}.

The ultimate goal of creating a character comprehension system is to equip a machine with practical capabilities that emulate what humans can accomplish. For example, understanding personae can facilitate
story memorization and generation of new statements consistent with the story~\cite{riedl2010narrative}. Such capabilities could be valuable in the construction of dialog engines that help people address practical problems such as those experienced in customer service encounters ~\cite{mairesse2007personage,zhang2018personalizing,urbanek2019learning}.
More importantly, understanding the persona of a particular person can help chatbots to understand the intention behind a human user’s  language~\cite{bender2020climbing}, which can lead to better services and ultimately give systems the ability to demonstrate behaviors that users interpret as empathetic. For instance, \emph{Amy}'s last sentence in Figure~\ref{fig:bookqa_example} is a joking braggadocio to remind her boyfriend to value her more.
% is a can be understood in multiple ways on its surface form. 
Only when \emph{Sheldon} understood
% \footnote{But he may not \Doge.} 
the facts of their relationship as a couple and \emph{Amy}’s temporary show-off mentality could he see her true intentions.\footnote{But he cannot \goutou.}

Despite the importance of this capability, there has been limited attention to modeling characters in stories in the natural language processing (NLP) community.\footnote{In contrast, plot comprehension is a popular NLP topic, especially on event structures~\cite{finlayson2012learning, elsner2012character,sims2019literary,lal2021tellmewhy,han2021ester}.}
Most existing character-centric prediction tasks have the input sources in expository text such as synopsis (summaries) of stories~\cite{brahman2021let} or non-narrative dialogues~\cite{zhang2018personalizing,urbanek2019learning,li2020aloha}.
A few exceptions work on stories, but focus on limited aspects of persona, such as facts for coreference resolution~\cite{chen2016character}, personality~\cite{bamman2013learning,flekova2015personality} and character relationships~\cite{iyyer2016feuding}, with only a few \citet{chen2016character,flekova2015personality} providing  evaluation benchmarks.
Besides the limited persona aspect coverage, these models also lack the ability to take into account a theory of mind (\textbf{ToM}) which is the knowledge of epistemic mental states that humans use to describe, predict, and explain behavior ~\cite{baron1997mindblindness}.

In this paper, we propose the first task focused on character comprehension in stories, to assess the ability of mental model construction in NLP. A character's words are her direct reflection of the contexts, conditioned on her character model~\cite{holtgraves2010social}. Our task, \textbf{\datasetname (TVSG)}, 
aims to guess anonymous speakers using dialogues, scene descriptions and historical scenes, which requires models to interpret the behavior of characters in the form of dialogues, which meets the requirements for the evaluation of ToMs.

Through experiments and human studies we found the following results: First, human performance was nearly perfect, while the model performed poorly. Second, although our TVSG has a simple task setup, it has a surprisingly \emph{wide coverage of persona understanding skills} including the linguistic styles, personality types, factoids, personal relations, and the memories of characters' previous experience. 
Third, most of the cases ($>$60\%) require \emph{identification and understanding of characters' historical experiences} to resolve. Among them, many rely on facts of characters that are not explicitly described in texts but need to be inferred from event history. The wide persona coverage and heavy dependency on  history present challenges to existing NLP techniques; and explain the more than 20\% accuracy gap between our baselines and humans.

We make the following contributions:

(1) We propose the research direction of character comprehension in stories; with an extended survey (Section~\ref{sec:related} and Appendix~\ref{app:survey}) discussing the differences and challenges compared to related work.

(2) We propose the first task and dataset for multi-aspect persona (especially ToM) understanding in stories (Section~\ref{sec:dataset}).

(3) We propose a new schema to analyze the required evidence for character understanding; and conduct human studies to analyze the required skills of our task (Section~\ref{sec:analysis} and Appendix~\ref{app:reasoning_type}).

(4) We propose new model architectures as the initial step of this direction; and conduct comprehensive experiments to provide insights to future work (Section~\ref{sec:method} and \ref{sec:exp}).

%%%%%%%%%%%%%%%%%%%%%%%%%%%%%%%%%%%%%%%%%%%%%%  
%%%%%%%%%%%%%%%  Related Work  %%%%%%%%%%%%%%%
%%%%%%%%%%%%%%%%%%%%%%%%%%%%%%%%%%%%%%%%%%%%%%

\section{Related Work}
\label{sec:related}

\begin{table*}[t]
\small
\centering
\resizebox{0.9\textwidth}{!}{%
\begin{tabular}{lcccccccccc}
\toprule
\multirow{2}{*}{\textbf{Dataset}}& \multirow{2}{*}{\textbf{Task Format}} & \multicolumn{2}{c}{\hspace{0.8 in}\textbf{Narrative Type}} & \multicolumn{3}{c}{\textbf{Assessed Narrative Comprehension Skills}} \\
& & \textbf{Source} & \textbf{Length} & \textbf{Plot Structures} & \textbf{Character Facts} & \textbf{Character ToMs} \\
\midrule
\textbf{MCTest} &  Multi-choice QA & {\begin{tabular}[c]{@{}c@{}}Short fiction \scriptsize{(Children stories)}\end{tabular}} & $\sim$20$^\ast$ & \checkmark &  &   \\
\textbf{BookTest} & Cloze test & {\begin{tabular}[c]{@{}c@{}}Literature \scriptsize{(Excerpt)}\end{tabular}} & -- & \checkmark & &  \\
\textbf{\cite{ma2018challenging}} & Cloze test &{\begin{tabular}[c]{@{}c@{}}TV show transcripts \scriptsize{(Scenes)}\end{tabular}} & $\sim$20 & \checkmark & \\
\textbf{NarrativeQA} & Generative QA & {\begin{tabular}[c]{@{}c@{}}Movie Scripts, Literature \scriptsize{(Full stories)}\end{tabular}} & $\sim$11K$^\ast$ & \checkmark & \checkmark & \\
\textbf{FriendsQA} & Extractive QA &{\begin{tabular}[c]{@{}c@{}}TV show transcripts \scriptsize{(Scenes)}\end{tabular}} & $\sim$20$^\ast$  & \checkmark & \checkmark\\
\textbf{NovelChapters/BookSum} &Summarization &{\begin{tabular}[c]{@{}c@{}}Literature \scriptsize{(Chapters or Full stories)}\end{tabular}} &$\sim$4K &\checkmark&&\\
\textbf{SummScreen} &Summarization &{\begin{tabular}[c]{@{}c@{}}TV show transcripts \scriptsize{(Scenes)}\end{tabular}} &$\sim$330&\checkmark&&\\
{\begin{tabular}[c]{@{}c@{}}\textbf{\cite{chen2016character} /} \\ \textbf{\cite{chen2017robust}}\end{tabular}} & Coref Resolution &{\begin{tabular}[c]{@{}c@{}}TV show transcripts \scriptsize{(Episodes or scenes)}\end{tabular}} & $\sim$20/260$^{\dagger}$ & \checkmark & \checkmark & \\
\textbf{\cite{flekova2015personality}} & Classification &{\begin{tabular}[c]{@{}c@{}}Literature \scriptsize{(Full stories)}\end{tabular}} & $\sim$22K  & &\checkmark & \\
\midrule
\textbf{\datasetnamens} & Multi-choice & {\begin{tabular}[c]{@{}c@{}}TV show transcripts \scriptsize{(Full stories)}\end{tabular}} &  $\sim$50K & \checkmark $^{\ddagger}$ & \checkmark & \checkmark \\
\bottomrule
\end{tabular}
}
\caption{\small{Properties of existing narrative comprehension datasets compared to \datasetnamens. 
* Numbers are not reported in the original paper so we calculated them from the dataset.
$\dagger$\cite{chen2017robust} proposes two settings: single scene and the whole episode.
$\ddagger$Our task requires reasoning based on history scenes, which is a form of plot understanding.}}
\vspace*{-3mm}
\label{tab:existing_ds}
\end{table*}

In this section we mainly discuss and compare related areas: the assessment benchmarks for general narrative comprehension skills; and the tasks specifically designed for character-centered predictions over narratives. Table \ref{tab:existing_ds} gives a summary of these narrative comprehension tasks, associated with their required comprehension skills. We also reviewed studies on character-centered tasks over non-narrative texts like synopses and chit-chat (\emph{i.e.}, not story-related) conversations. Detailed rationales of the required skills for each task are discussed in Appendix~\ref{app:survey}.

\medskip
\noindent\textbf{Assessment of Narrative Comprehension }
There are many forms of reading comprehension tasks such as cloze tests~\cite{bajgar2016embracing,ma2018challenging}, question answering~\cite{richardson2013mctest,kovcisky2018narrativeqa,yang2019friendsqa,lal2021tellmewhy,xu2022fantastic}, and text summarization~\cite{ladhak2020exploring,kryscinski2021booksum,chen2021summscreen}.
Most of these tasks are built on very short stories or can be solved in segments of a story, and therefore present limited challenges to understanding the elements of the story, especially the characters.
The exceptions are NarrativeQA~\cite{kovcisky2018narrativeqa} and three summarization tasks which are mainly event-centric tasks focusing on understanding the plot structures in the stories.
The NarrativeQA consists of a small portion of character-related questions according to the human study in~\cite{mou2021narrative}, but mainly about simple facts of characters like age, place of birth and profession.
Finally, text games~\cite{hausknecht2019interactive} have been proposed as a reinforcement learning task that requires understanding of narrative fiction stories. Studies have been conducted~\cite{guo2020interactive,yao2021reading} to investigate the roles reading comprehension plays in these games.

\medskip
\noindent\textbf{Character-Centric Prediction over Narratives}
The task of coreference resolution of story characters~\cite{chen2016character,chen2017reading} is most closely related to our \datasetnamens.
These tasks focus on identifying the characters mentioned in multiparty conversations,  
which mainly requires the understanding of discourse relations and assessment of personal facts. However, coreference does not assess the modeling of the character's theory-of-mind, especially the character’s memories, as there are no predictions of character behaviors involved. The prediction of fictional characters' personality types by reading the original stories~\cite{flekova2015personality} is another character-centric task related to the present work.
This work covers only the character's personality such as the big five and the MBTI types,
also a perspective of the persona our work considers.

\medskip
\noindent\textbf{Character-Centric Prediction over Non-Narratives }
Many tasks do not use the original story, but rather a summary of it. For example, the textual entailment task LiSCU~\cite{brahman2021let} links an anonymous character summary to the name appearing in the story's summary. The usage of summaries precludes ToM modeling, as discussed in Appendix~\ref{app:synopsis_clarify}.
Personalized dialogue generation~\cite{mairesse2007personage,walker2012annotated,zhang2018personalizing,urbanek2019learning,li2020aloha} benchmarks are based on daily chit-chats. They usually cover a single facet of a multi-dimensional persona~\cite{moore2017five}, \emph{e.g.}, personal facts~\cite{zhang2018personalizing} or personality types~\cite{mairesse2007personage,li2020aloha}. The LIGHT environment~\cite{urbanek2019learning} covers both facts and personalities. None of the above covers a comprehensive persona like ours, especially how a character’s past experience builds her ToM.

Authorship attribution has a parallel goal to ours, insofar as it aims at guessing author identities from the texts they wrote~\cite{ni2019justifying,andrews2019learning,bevendorff2020overview}. 
These tasks differ from ours in two respects: first, multiple prose examples generated by the same author do not usually form consecutive plot lines, and second, they rarely model the event history of depicted characters.
On this basis, they mainly require understanding authors' writing styles rather than building mental models of facts, events, and experiences.

\section{Our \datasetname Benchmark}
\label{sec:dataset}

\subsection{Task Definition}

TVSG adopts a multi-choice setting. The goal is to guess the anonymous speakers who are the main characters (maximum number of 6 for each show) in the scene.
The models are provided with an anonymous scene that consists of $n$ lines $\tilde{\mathcal{S}}^{(t)}=\{\tilde{s}^{(t)}_1, \tilde{s}^{(t)}_2, ..., \tilde{s}^{(t)}_n\}$ ($t$ stands for the $t$-th scene in the entire show).
Each line $\tilde{s}_i$ can be either a dialogue turn or a background description.
When the line is a dialogue turn, it is associated with an anonymous speaker ID (with the form of $\text{P}_x$, $1 \le x \le 6$) of a main character, or the real name of a supporting character.
Similarly, we introduce the notation of the standard scene $\mathcal{S}^{(t)}=\{s^{(t)}_1, s^{(t)}_2, ..., s^{(t)}_n\}$, which has the same definition as the anonymous scenes, with the only difference that the dialogue turns always have their real names of speakers associated.

The anonymous scene $\tilde{\mathcal{S}}^{(t)}$ is associated with a candidate set $\mathcal{C}^{(t)}= {c^{(t)}_1,...,c^{(t)}_k}$, $k \le 6$, with each character $c^{(t)}_j$ is a main character who appears in $\mathcal{S}$. The goal is thus predicting each $\text{P}_x$'s actual role $c^{(t)}_j$, \emph{i.e.}, a match $\pi(\cdot)$ from the anonymous IDs to the real characters, conditioned on the scene $\tilde{\mathcal{S}}^{(t)}$ and all  previous scenes $S^{(1:t-1)}$:
\vspace{-2mm}
\begin{equation}
\small
\begin{aligned}
    P(\text{P}_x = c^{(t)}_j\vert \tilde{\mathcal{S}}^{(t)}, S^{(1:t-1)}). \label{eq:problem}
\end{aligned}
\end{equation}

\subsection{Dataset Collection}
We use community contributed transcripts from The TV MegaSite (TMS)\footnote{\url{http://tvmegasite.net}.} like~\cite{chen2021summscreen}. Our scenes are from the scripts of five popular TV series: \emph{Friends}, \emph{The Big Bang Theory (TBBT)}, \emph{The Office}, \emph{Frasier} and \emph{Gilmore Girls}.

\begin{table*}[t!]%[width=\textwidth]
    \small
    \centering
    \renewcommand{\arraystretch}{1} % Default value: 1
    \resizebox{0.9\textwidth}{!}{
    \begin{tabular}{lc cc cc cc cc} 
        \toprule
        \multirow{2}{*}{\bf Show} & \multirow{2}{*}{\bf train} & \multirow{2}{*}{\bf dev} & \multirow{2}{*}{\bf test} & \multicolumn{2}{c}{\bf \#tokens per utterance} & \multicolumn{2}{c}{\bf \#tokens per scene} & \multicolumn{2}{c}{\bf\#tokens per character}\\ 
        \cmidrule(lr){5-6}
        \cmidrule(lr){7-8}
        \cmidrule(lr){9-10}
        &&&&\hspace{0.1 in}{\bf avg}&\bf max&\hspace{0.1 in}{\bf avg}&\bf max&\hspace{0.1 in}{\bf avg}&\bf max\\
        \midrule
        \texttt{Friends} &2,418 &210&211&\hspace{0.1 in}21&350&\hspace{0.1 in}862&6,817&190,932&516,191\\
        \texttt{TBBT} &1,791 &130 &130&\hspace{0.1 in}19&364&\hspace{0.1 in}414&6,051&167,027&183,748 \\
        \texttt{Frasier} &1,368&140&141&\hspace{0.1 in}16&363&\hspace{0.1 in}812&14,276&165,483&475,372\\
        \texttt{Gilmore\_Girls} &1,495&141&142&\hspace{0.1 in}19&336&\hspace{0.1 in}360&4,572&105,723&214,779 	 \\
        \texttt{The\_Office} &3,699&198&199&\hspace{0.1 in}19&338&\hspace{0.1 in}123&1,660&58,676&132,992 \\
        \midrule
        total&10,771&819&823&\hspace{0.1 in}18&364 &\hspace{0.1 in}371&14,276&137,568&516,191 \\
        \bottomrule
    \end{tabular}}
    \vspace{-2mm}
    \caption{\small{Statistics of our \datasetnamens. The numbers in the first 3 columns refer to the total numbers of scenes.}}
    \vspace{-3mm}
    \label{tab:dataset_stats}
    
\end{table*}

\medskip
\noindent\textbf{Data Cleaning }
Our data consists of character dialogues and background descriptions. The dialogues start with the characters' names. One or more turns of dialogue between characters comprise a scene. Scenes are separated by short background cus that begin with markers such as location (e.g. ``\emph{Howard's car}'', ``\emph{Kingman Police Station}''), special words (e.g., ``\emph{Scene}'', ``\emph{Cut}''), or symbols (e.g. ``\emph{[ ]}''). We created a rule-based parser which splits the content of an episode into multiple independent scenes using scene separation markers.

\medskip
\noindent\textbf{Character Recognition and Anonymization }
We used the main characters' names to identify their dialogues within each scene and randomly labeled them with speaker IDs (i.e., P0, P1). Since the different names of the characters, such as nicknames, first names, and last names, mark the dialogue in a mixed way, in order to match the lines to the correct speaker, we first identified the main characters in each TV series by consulting Fandom's cast list. Then, we calculated the frequency of speech to find references to the same main character's name.

%%%%%%%%%%%%%%%%%%%%%%%%%%%%%%%%%%%%%%%%%%%%%%  
%%%%%%%%       Human Study       %%%%%%%%%%%%%
%%%%%%%%%%%%%%%%%%%%%%%%%%%%%%%%%%%%%%%%%%%%%%

\section{Analysis of Our Benchmark}
\label{sec:analysis}
We propose a comprehensive \textbf{schema of persona types} for the machine narrative comprehension. The schema facilitates the analysis of the challenges in our task; and provides insights into the deficiencies in current narrative comprehension models, by allowing a decomposition of model performance to the dimensions of categories (Section~\ref{sec:exp}).

\subsection{Our Annotation Schema for the Human Study}
Two researchers with backgrounds in psychology, linguistics, NLP, and education developed and tested an inductive coding method based on the methods of grounded theory~\cite{glaser2017discovery}. They conducted three rounds of independent annotation and discussion of the evidence needed to identify the characters, using 10 randomly selected scenes for each round. After each discussion, they updated the codebook accordingly. Modifications to the codebook led to the achievement of saturation during the process. Then the two researchers coded a total of 318 characters from 105 scenes of Friends and The Big Bang Theory. The annotation interface appears in Appendix~\ref{app:annotation}.

This schema 
\textbf{categorizes the required evidence to resolve the task} into four persona data types: 
\emph{linguistic style}, \emph{personality}, \emph{fact}, \emph{memory}. 
Table~\ref{tab:annotation_agreement} reports inter-rater reliability calculated by Cohen’s Kappa~\cite{cohen1960coefficient}. The kappa values range from 0.76 to 0.87 and would all be considered satisfactory~\cite{viera2005understanding}, reflecting the success of our codebook and process.

We have one additional type, \emph{inside-scene}, referring to tasks that can be resolved purely within local contexts, thus not requiring persona understanding. To better depict how these pieces of evidence are used in human rationales, we added two complementary categories: (1) how the task instance \textbf{relies on the history scenes}, and  
(2) when there are multiple pieces of evidence required, what \textbf{types of reasoning skills} are used to derive the answer from the evidence (see Section~\ref{app:reasoning_type}). Table~\ref{fig:evidence_category} shows the definitions of each evidence type. We provide examples of each evidence type in Section~\ref{app:evidence_type}.

\subsubsection{Major Evidence Types}
\label{ssec:evidence_type}
\medskip
\noindent\textbf{Linguistic Style } Personalized language patterns that reflect individual differences in self expression and is consistently reliable over time and situations~\cite{pennebaker1999linguistic}.

\medskip
\noindent\textbf{Personality} Stable individual characteristics~\cite{vinciarelli2014survey} that can distinguish ``internal properties of the person from overt behaviors''~\cite{matthews2003personality}.

\medskip
\noindent\textbf{Memory}
A character's episodic memory of events from previous episodes and the semantic memory\footnote{Semantic memory is the characters' general world knowledge that they accumulate over time~\cite{reisberg2013oxford}. Episodic memory, on the other hand, is the characters' memory of specific experiences in their lives~\cite{tulving2002episodic}} inferred from events.
Note here we want to model the memory of a particular character, \emph{i.e.}, the historical scenes experienced by the particular character instead of from the audiences' perspective.
A character's memory is crucial for humans to build her ToM, but is largely ignored as a part of persona in previous research.

\medskip
\noindent\textbf{Fact } The truth about characters as opposed to interpretation, which can usually be represented as knowledge triples. 
\begin{itemize}[leftmargin=*]
\vspace{-0.5em}
\setlength\itemsep{0em}
    \item\textbf{Attribute } All explicitly provided factual character identity information in the TV series setting, such as race, occupation, and education level.

    \item\textbf{Relationship } Relationship includes social relationships (e.g., husband and wife) and dramatic relationships (e.g., arch-enemy). When talking to people with different relationships, characters change their identity masks by using different words~\cite{gergen1972multiple}. 
    
    \item\textbf{Status } The emotional or psychological status of a character when facing a specific situation. 
\end{itemize}

\medskip
\noindent\textbf{Inside-Scene } The textual evidence inside the scene, independent from the character's persona. 
\begin{itemize}[leftmargin=*]
\setlength\itemsep{0em}
% \vspace{-1em}
    \item \textbf{Background } Background introduction and descriptions in other character dialogues.
    
    \item \textbf{Mention } The character's name or alias is called out by other characters. Although mention is persona-independent, it still presents challenging cases. In a multi-person, multi-round dialog, because anaphora in the current sentence may not be bound to an antecedent on the right frontier of the structure,
common sense analysis of conversational coherence is needed to determine which speaker is being referred to.
\end{itemize}
% \vspace{-8mm}
\medskip
\noindent\textbf{Exclusion } 
A guessing technique for elimination using a given list of characters which is neither evidence nor inference, but it depends on the character list provided within the scene, so we include it as a subcategory of inside-scene evidence.

\subsubsection{Dependence on History}
\label{ssec:dependence_type}
To understand how much humans rely on memory to identify a character, the annotators coded whether the evidence necessary to solve the task depends directly on historical events or whether it depends indirectly on history by abstracting from historical events.

\medskip
\noindent\textbf{Direct Dependency} Characters that can only be identified through events that are explicitly expressed in previous episodes,\footnote{
If a character can be identified with evidence of both \emph{Memory} and \emph{Inside-Scene}, it will be labeled as \emph{No-Dependency}.} for example:

    {\small
    \centering
    \begin{tabular}{p{7.3cm}}
      \rowcolor{Gray} {\textbf{Background:}} (from \texttt{TBBT}) \textit{[The stairwell]}\\
      \rowcolor{Gray} {\textbf{Candidates:}}  \textit{\{Leonard, Penny\}}\\
      \rowcolor{Gray} 
      \textbf{\textcolor{coutput}{P0}:} \textit{There's something I wanted to run past you.}\\
      \rowcolor{Gray} \textbf{\textcolor{cinput}{P1}:} \textit{What's up?}\\
      \rowcolor{Gray} \textbf{\textcolor{coutput}{P0}:} \textit{Mm, the guys and I were thinking about investing in Stuart's comic book store. Is that okay?} \\
      \rowcolor{Gray} \textbf{\textcolor{cinput}{P1}:} \textit{Why are you asking me?} \\
      \rowcolor{Gray} \textbf{Answer:} \textbf{\textcolor{coutput}{P0}} $\rightarrow$ \emph{Leonard}\\ 
      \rowcolor{Gray} \textbf{Rationale:} In a previous scene, Leonard and his friends discussed about investing in Stuart's store, so he is the only one between the two who has this memory.
      \end{tabular}\\}

\medskip
\noindent\textbf{Indirect Dependency} Characters can only be identified using evidence not explicitly expressed in previous episodes, but inducible from previous events. For example, 
\emph{Personality} can be inferred from the character's previous behavior.\footnote{The annotation of indirect dependency is very subjective as different annotators may have memory of previous scenes and use different evidence to guess the character.}

    {\small
    \centering
    \begin{tabular}{p{7.3cm}}
      \rowcolor{Gray} {\textbf{Background:}} (from \texttt{Friends}) \textit{[Central Perk]}\\
      \rowcolor{Gray} {\textbf{Candidates:}}  \textit{\{Joey, Rachel, Ross\}}\\
      \rowcolor{Gray} 
      \textbf{\textcolor{coutput}{P0}:} \textit{Here you are (Hands Rachel a cup of coffee)}\\
      \rowcolor{Gray} \textbf{\textcolor{cinput}{P1}:} \textit{Thank you Joey. You know what? I'm not even sure I can have caffeine.}\\
      \rowcolor{Gray} \textbf{\textcolor{cgroundtruth}{P2}:} \textit{I went thru this with Ben and Carol. One cup of coffee won't affect your milk.} \\
      \rowcolor{Gray} \textbf{\textcolor{cinput}{P1}:} \textit{Yeah. Just to be sure I'm gonna call Dr. Wiener.} \\
      \rowcolor{Gray} \textbf{Answer:} \textbf{\textcolor{cgroundtruth}{P2}} $\rightarrow$ \emph{Ross}\\ 
      \rowcolor{Gray} \textbf{Rationale:} There is no actual scene with Ross going through this with Carol; the answer is inferred based on Ross' known relations to Ben (parent-child) and Carol (ex-spouse). Thus the evidence about Ross  has indirect dependency on scene history.
      \end{tabular}\\}

If the answer can be inferred within the scene, like answering \textbf{\textcolor{coutput}{P0}} $\rightarrow$ \emph{Joey} in the above example.
We have a special rule on the \emph{Exclusion} evidence type -- If a character can only be inferred on the basis of other characters being eliminated, it should have dependency type labeled if any of the other character has a history dependency. 
In other words, when guessing the identity with \emph{Exclusion} requires history dependency on another character, the dependency type is transitive.

\begin{table}[t]
\scriptsize
\centering
% \small
% \resizebox{0.45\textwidth}{
\begin{tabular}{lp{21mm}cc}
\toprule
& Evidence Type & Friends(\%) & TBBT(\%) \\ \midrule
\multirow{9}{*}{\rotatebox[origin=c]{90}{(a)}} 
        &Ling. Style            & 0.66    & 9.93      \\
        &Personality            & 7.28    & 21.85     \\   
        &Fact        &   20.53  & 33.12      \\   
        &\quad (Attribute)        & 2.65    & 8.61      \\       
        &\quad (Relation)         & 16.56   & 22.52     \\   
        &\quad (Status)           & 1.32    & 1.99      \\   
        &Memory                 & 36.42   & 27.15     \\   
        &Inside-Background     & 33.11   & 12.58     \\       
        &Inside-Mention        & 15.23   & 15.23     \\       
        &Exclusion      & 8.61    & 22.52     \\ 
\midrule
\end{tabular}

\begin{tabular}{llccc} 
% \midrule \midrule
\midrule
& Dependence of Hist. & Friends(\%) & TBBT(\%)         \\ 
\midrule    \multirow{3}{*}{\rotatebox[origin=c]{90}{(b)}}
        & No Dep.               & 53.64     & 32.45     \\
        & Direct Dep.           & 26.49     & 36.42     \\
        & Indirect Dep.         & 19.87     & 31.13     \\
\bottomrule
\end{tabular}
\vspace{-2mm}
\caption{\small{Percentage of the required evidence types in the two TV shows, \texttt{Friends} and \texttt{The Big Bang Theory}.}}
\label{tab:percentage_evidence}
% \vspace{-4mm}
\end{table}

\subsection{Analysis}
\medskip
\noindent\textbf{Main Statistics} Table \ref{tab:percentage_evidence} shows the proportions of the required evidence types and dependency of history. According to the statistics, history is an important factor in guessing the characters. 46.36\% of the examples from Friends and 67.55\% examples from the Big Bang Theory require history.

\medskip
\noindent\textbf{Human Performance Accuracy }
One annotator (who had not watched the seasons under evaluation) obtained nearly perfect accuracy in guessing the characters in \texttt{FRIENDS} (98.68\%), and a lower but still good accuracy in \texttt{TBBT} (89.82\%). A second annotator (who had watched all episodes and thus would be considered an expert) confirmed that most the error cases were unsolvable given only data extracted from  scenes. We list the unsolvable cases and human mistakes in Appendix~\ref{app:human_error}.

\begin{table}[t!]
    \small
    % \fontsize{9}{10}\selectfont
    \centering
    \begin{tabular}{lcc}
        \toprule
        \textbf{Category}   &\bf $\mathbf{\kappa}$(\%)  \\
        \midrule
        Evidence type\\
        \quad Coarse-grained types       &  81.53\\
        \quad Fine-grained types & 80.99\\
        \midrule
        Dependence of history\\
        \quad Direct dependence only & 82.02 \\
        \quad All dependency types & 75.51\\
        \midrule
        \rowcolor{purple!40} Reasoning Type$^{\dagger}$ &   87.21 \\
        \bottomrule
    \end{tabular}
    \vspace*{-2mm}
    \caption{\small{Annotation agreement. $\dagger$: see our extended study in Appendix~\ref{app:reasoning_type}. We list the number for reference.}}
    \vspace*{-2mm}
    \label{tab:annotation_agreement}
\end{table}

\begin{figure}[t!]
    % \fontsize{14}{10}\selectfont
    \centering
    \includegraphics[width=0.45\textwidth]{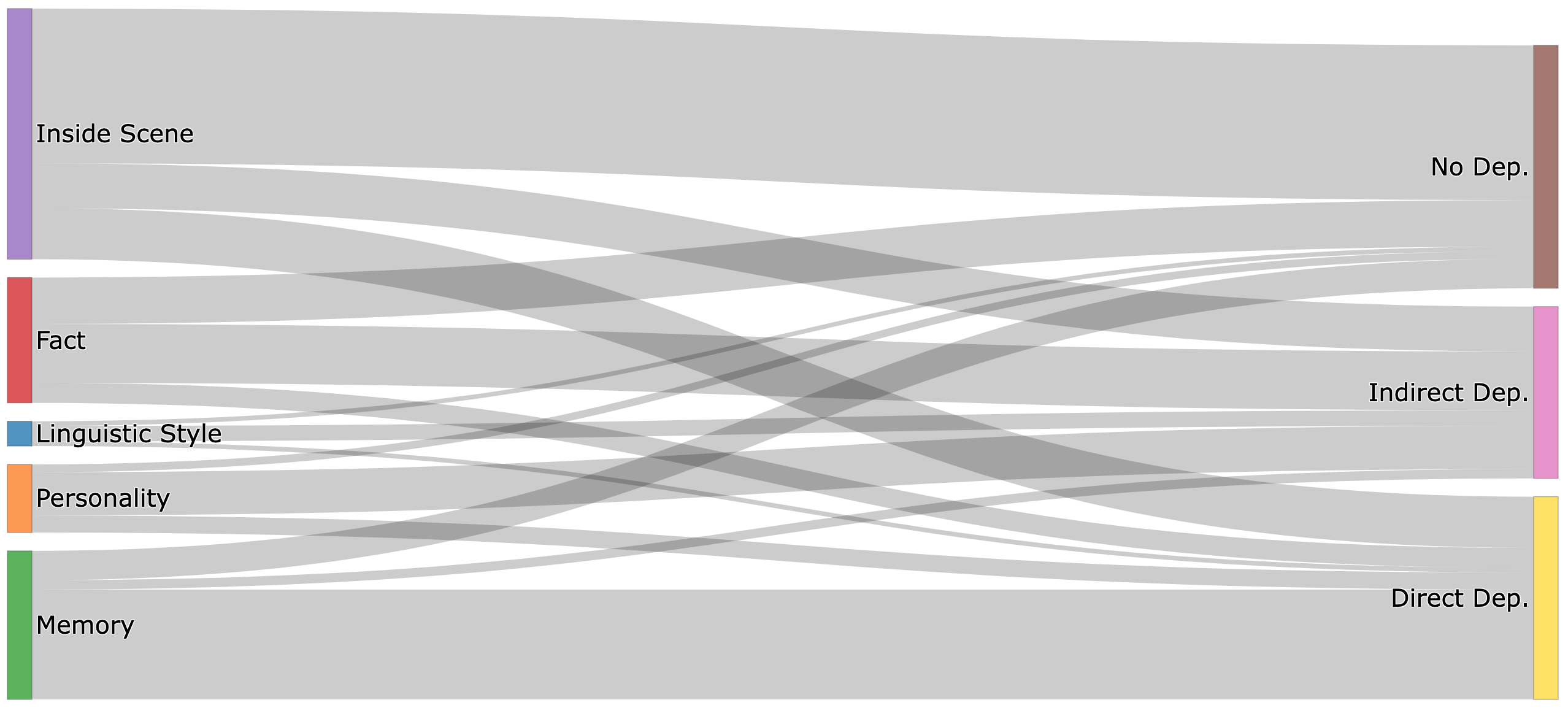}
    \caption{\small{Visualization of the flow from the required evidence types to their dependence of history.}}
    \label{fig:annotation_category_flow}
    \vspace{-3mm}
\end{figure}

\medskip
\noindent\textbf{Correlation between Evidence Types and History Dependence}
Figure~\ref{fig:annotation_category_flow} visualizes the flow from evidence types to the dependency of history. Most of them are correlated. Personality and history dependency are most closely related.

\section{Methods}
\label{sec:method}
\begin{figure*}
\centering
\includegraphics[width=0.95\textwidth]{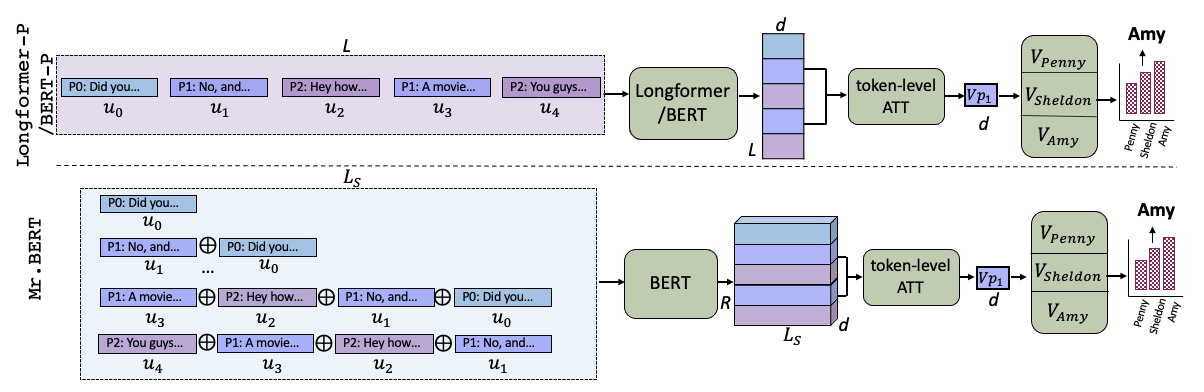}
\vspace{-2mm}
\caption{\small{Our two proposed model architectures for the character prediction task.}}
\centering
\label{fig:MR_BERT}
\end{figure*}

Inspired by the successes of pre-trained Transformers on reading comprehension tasks, we benchmarked our \datasetname by building baseline solutions on top of these pre-trained models. The key challenge of our task is that the prediction relies on how a character reacts to the scenario with her/his words, therefore the embedding of each utterance should be highly \textbf{context-aware}.
This requires handling a whole scene as input, which is usually over the limits of these pre-trained models.

We propose two solutions. The first is to benefit from sparse attention, specifically, Longformer~\cite{beltagy2020longformer}. 
The second is to organize each utterance with its necessary history context as one \textbf{row}, and have a BERT model to encode each relatively short row independently.
For both models, we finally conduct attentive pooling for each character over the contextualized embeddings of all her utterances for prediction.

Our baselines simplified the problem by (1) ignoring the historical scenes in Eq. \ref{eq:problem}; and (2) making independent prediction of characters within a scene. The former poses the challenge of handling longer contexts and the latter requires specific predictor design. We believe both are important to handle in future work.

\setlength{\abovedisplayskip}{3pt}
\setlength{\belowdisplayskip}{3pt}
\subsection{Transformers with Character-Pooling}
Our first approach (the top in Figure~\ref{fig:MR_BERT}) is denoted as Longformer-Pooling (or \textbf{Longformer-P}).

\medskip
\noindent\textbf{Scene Encoding}
The input $\tilde{S}$ to the model includes the concatenation of all the utterances in an anonymous scene.
Each utterance is prefixed by a speaker ID token and suffixed by a separation token, \emph{i.e.},

\vspace{-4mm}
\begin{equation}
\small
    T_i = \text{[P}_{x_i}\text{]} \oplus U_i \oplus \text{[SPLIT]}   \label{eq:utterance}
\end{equation}
\vspace{-5mm}
\begin{equation}
\small
    \tilde{S} = T_0 \oplus T_1 \oplus ... \oplus T_N,      
\end{equation}

where $U_i$ is the $i$-th utterance and $\text{[P}_{x_i}\text{]}$ is its speaker ID (e.g., $\text{[P}_0\text{]}$ and $\text{[P}_1\text{]}$). [SPLIT] is a special token. $\oplus$ denotes concatenation.
We use a Longformer to encode the whole $\tilde{S}$, to make the embedding of each utterance token \emph{context-aware}, \emph{i.e.}, $\mathbf{H} = \textrm{Longformer}(\tilde{S}) \in \mathbb{R}^{L \times D}$.

\medskip
\noindent\textbf{Character-Specific Attentive Pooling}
For each character ID $\text{P}_{x}$, we have a token-level mask $M_x \in \mathbb{R}^{L \times 1}$ such that $M_x[j]$$=$$1$ if the $j$-th word belongs to an utterance of $\text{P}_{x}$ and $M_x[j]$$=$$0$ otherwise. For each $\text{P}_{x}$, we then collect the useful information from all her utterances selected by $M_x$ as:
\begin{equation}
\small
\begin{aligned}
    A &= \textrm{Attention}(\mathbf{H}) \\
    \alpha_x &= \textrm{Softmax}(A \odot M_x)   \label{eq:attn_pooling_2},
\end{aligned}
\end{equation}

where $\textrm{Attention}(\cdot)$ is a one-layer feedforward network to compute the token-level attention weights. The character-specific attention $\alpha_x$ is then used to pool the hidden states to summarize a character representation in the input scene $\tilde{S}$ and make the prediction: $P(\text{P}_x=c|\tilde{S}) =  f_{k}(\mathbf{H}^T \alpha_x )$.
Here $f_{k}: \mathbb{R}^{d \times 1} \xrightarrow{} \mathbb{R}^{C \times 1}$ is the character classifier for the $k$-th TV show.

\subsection{Multi-Row BERT}
The second approach (the bottom in Figure~\ref{fig:MR_BERT}) is denoted as the multi-row BERT (\textbf{MR. BERT}).
We split the long scene $\tilde{S}$ into multiple segments $\{\tilde{s}_i\}$. Encoding the segments reduces the overall complexity from $O(L^2)$ to $O(RL_s^2)$, where $R$ is the number of segments and $L_s \ll L$ is the maximum segment length. 
To construct each segment $E_i$, we take an utterance $T_i$ as in Eq.~\ref{eq:utterance} and concatenate it with the nearest history utterances $T_{i'} (i'<i)$ until arriving the maximum length $L_s$. The $R$ segments for each instance yield $\{\tilde{s}_i\}$ as follows:
\begin{equation}
\small
\begin{aligned}
    E_i &= T_{t_i} \oplus \text{[SEP]} \oplus T_{t_i-1} \oplus T_{t_i-2} \cdots  \label{eq:multirow_segment}\\
    \{\tilde{s}_i\} &= [E_1; E_2; \cdots; E_R].
\end{aligned}
\end{equation}

Then we encode the $\{\tilde{s}_i\}$ with a BERT encoder:
\begin{equation}
\small
\begin{aligned}
    \mathbf{H} &= \textrm{BERT}(\{\tilde{s}_i\}) \in \mathbb{R}^{R \times L_s \times D}.
\end{aligned}
\end{equation}

%%%% Version Camera-Ready %%%%
Different from Longformer-P, we have a segment-level mask $M_x \in \mathbb{R}^R$ for each character ID such that  $M_x[j] = 1$ if the first utterance in the $j$-th row (i.e., $T_{t_j}$ in Eq.~\ref{eq:multirow_segment}) is said by $\text{P}_x$.
Applying the same attentive pooling technique to each segment following Eq.~\ref{eq:attn_pooling_2}, we obtain $R$ segment embeddings $\{\mathbf{E}_i\}_1^R$. We take the concatenation of these embeddings as the new input to the show-specific predictor and calculate the probability distribution of $\text{P}_x$ being each character, \emph{i.e.},
% \vspace{-2mm}
\begin{equation}
\small
\begin{aligned}
    P(\text{P}_x=c|\tilde{S}) =  f_{k}([\mathbf{E}_1;\mathbf{E}_2;\cdots;\mathbf{E}_R]).
\end{aligned}
\end{equation}

Compared to Longformer-P, the MR. BERT model takes a smaller number of $R$ utterances and benefits from their concatenated contextual utterances.
To make the selection of $R$ utterances representative, we applies two tricks: (1) \emph{fill-empty}, which makes sure each $\text{P}_x$ has at least one segment selected; (2) \emph{the reverse trick}, which selects the utterances starting from the end of scene to the start -- as the utterances at the end have more histories, they cover more contents from the scene if selected.

\section{Experiments}
\label{sec:exp}

\begin{table*}[t!]%[width=\textwidth]
    \small
    \centering
    \renewcommand{\arraystretch}{1} % Default value: 1
    \resizebox{\textwidth}{!}{
    \begin{tabular}{lcccccccccccc} 
        \toprule
        \multirow{2}{*}{\bf System} & \multicolumn{2}{c}{\bf FRIENDS}& \multicolumn{2}{c}{\bf TBBT} & \multicolumn{2}{c}{\bf Frasier} &	\multicolumn{2}{c}{\bf Gilmore\_Girls}	& \multicolumn{2}{c}{\bf The\_Office}& \multicolumn{2}{c}{\bf Overall}\\
         & \bf dev & \bf test & \bf dev & \bf test & \bf dev & \bf test & \bf dev & \bf test & \bf dev & \bf test & \bf dev & \bf test \\ 
        \midrule
        Random & 35.23 & 31.59 & 33.08 & 37.79 & 34.74 & 31.61 & 36.43 & 38.90 & 44.30 & 46.71 & 36.79 & 36.59\\
        Vanilla Longformer & 67.79&60.63&61.58&63.95&85.11&82.06&79.84&74.52&70.92&71.60&72.55&69.72 \\
        \quad repl with BERT & 65.60&59.58&61.58&58.43&85.11&84.30&81.91&70.41&67.56&68.54&71.65&67.76 \\
        \midrule
        % \cmidrule{2-4}
        Our MR. BERT & \bf 77.01&\bf 73.20&62.60&62.50&90.07&82.51&\bf 83.98&\bf 78.63&70.92&74.41&76.82&74.52 \\
        \quad - context & 62.92&57.19&59.54&63.95&81.64&76.23&74.42&67.12&66.00&67.37&68.33&65.54\\
        \quad - reverse trick &70.81&68.71&52.42&59.01&79.40&81.39&78.04&73.97&66.22&68.31&69.45&70.52\\
        \quad - fill-empty trick& 74.33&68.56&58.27&63.37&86.10&78.48&72.87&69.86&68.90&73.71&72.28&70.92\\
        \midrule
        Our Longformer-P & \bf 77.01&69.91&\bf 63.87&\bf 66.57&\bf 90.32&\bf 87.67&82.17&75.07&\bf 71.81&\bf 76.29&\bf 76.95&\bf 74.97\\ 
        % \midrule
        \quad maxlen=1000 & 74.16&66.77&63.36&64.24&86.10&85.65&79.33&72.05&73.83&76.06&75.25&72.74\\ 
        \quad repl with BERT & 68.12&58.83&61.32&63.95&82.63&76.91&68.48&65.75&72.48&71.83&70.49&66.79\\
        \midrule
        Human$^{*}$ & 98.68 & -- & 89.82 & --& --& --& --& --& --& -- \\
        \bottomrule
    \end{tabular}}
    \vspace*{-2mm}
    \caption{\small{Overall
    performance (\%) on our \datasetname task. 
    (*) Human evaluation was conducted on a subset of the dataset.}}
    \label{tab:overall_performance}
    \vspace{-5mm}
\end{table*}

We mainly evaluate the instance-level accuracy. An instance refers to a masked speaker in a scene, as defined in Eq. \ref{eq:problem}.

\subsection{Baselines and Implementation Details}

We also compare with the vanilla pre-trained Transformer baseline, \textbf{Vanilla Longformer Classifier}. The model conducts direct classification over the concatenation of a character's utterances in the scene. It can be viewed as a discriminative language model of the characters' lines.

We include the implementation details of the baseline and our models in Appendix~\ref{app:checklist}.

\subsection{Results}
\medskip
\noindent\textbf{Overall Results}
Table~\ref{tab:overall_performance} compares different models on our \datasetnamens. 
The proposed architectures 
beat our vanilla character classifier 
with large margins (4-5\%). 
However, human performance is significantly (21-26\%) better than the best models, 
showing models are still far from reaching human level of character understanding.

Among all the shows, \texttt{TBBT} is the most challenging one, while \texttt{Frasier} and \texttt{Gilmore Girls} are relatively simpler.
Given that there is no correlation between performance and scene lengths (Table~\ref{tab:dataset_stats}), this shows that the difficulty of the task mainly comes from the persona modeling, inference and reasoning.
Specifically, the \emph{Inside-Scene} evidence requires less persona understanding.
Therefore, the relatively smaller amount of \emph{Inside-Scene} cases makes \texttt{TBBT} more difficult.
Also the existing models are not good at resolving the related memory or facts from the history, thus the high ratio of \emph{history dependent} cases in \texttt{TBBT} also leads to lower performance.

\subsection{Analysis}
\begin{figure}[t!]
    % \fontsize{14}{10}\selectfont
    \centering
    \includegraphics[width=0.4\textwidth]{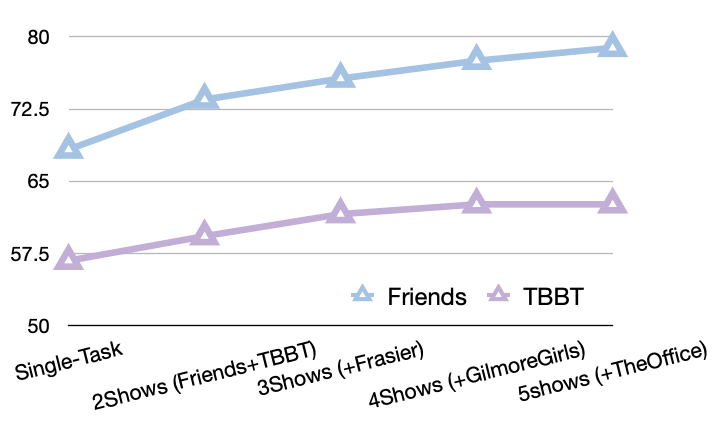}
    % \vspace{-3mm}
    \caption{\small{Learning curves of Friends and The Big Bang Theories with increasing training data from other shows.}}
    \label{fig:learning_curve}
\vspace{-2mm}
\end{figure}

\medskip
\noindent\textbf{Learning Curves}
We plots the learning curves of \texttt{Friends} and \texttt{TBBT}, with increasing number of shows used as training data (Figure~\ref{fig:learning_curve}). 
The curves become flat with all shows added, showing that our task has sufficiently data for training.

\medskip
\noindent\textbf{Scene-Level Performance}
Besides the instance-level accuracy, we further investigated the scene-level performance, \emph{i.e.}, the macro-average of instance-level accuracy in all the scenes. Table~\ref{tab:scene_accuracy} shows the results, together with the decomposed results on scenes containing different numbers of speakers. 
The results show the more characters involved, the lower the accuracy is, even though our model is making independent predictions of each speaker. One possibility is that there is fewer available utterances per speaker. In addition, a larger set of speakers may make the logical structure of the conversation more complex.

\begin{table}[!t]
    \small
    \centering
    \resizebox{0.45\textwidth}{!}{
    \begin{tabular}{lcccccc}
    \toprule
        & &\multicolumn{5}{c}{\bf \#Speakers Contained} \\ 
        & Overall & 2 & 3 & 4 & 5 & 6 \\ 
        \midrule
        FRIENDS & 80.6 & 86.5 & 80.8 & 66.3 & 75.0 & 56.7   \\ 
        TBBT &  67.7  & 77.0 & 66.7 & 57.0 & 55.0 & 47.9   \\ 
        \bottomrule
    \end{tabular}
    }
    \vspace*{-2mm}
    \caption{\small{Scene-Level accuracy decomposition.}}
    \label{tab:scene_accuracy}
% \vspace{-5mm}
\end{table}

\medskip
\noindent\textbf{Character-Level Performance}
Next, we examined whether our task is uniformly challenging for different characters, or whether there were certain characters that were more difficult to guess. Table \ref{tab:char_accuracy} shows the results, where the characters are ranked by the accuracy of their guesses. There are clear discrepancies in accuracy by character.

\begin{table}[!t]
    \small
    \centering
    \resizebox{0.45\textwidth}{!}{
    \begin{tabular}{lcc lcc}
    \toprule
        \multicolumn{3}{c}{\bf FRIENDS} & \multicolumn{3}{c}{\bf TBBT} \\ 
        \cmidrule(lr){1-3}
        \cmidrule(lr){4-6}
         & \#Utterance & Acc &  & \#Utterance & Acc \\ 
        \midrule
        rachel& 7,542 & 88.3 & sheldon & 8,131 & 87.0  \\ 
        joey & 6,550& 84.5   & penny & 5,314& 75.8  \\ 
        phoebe & 5,964 & 83.1   & leonard & 7,105 & 75.3 \\ 
        chandler & 6,804& 71.2 & raj & 3,033 & 52.5 \\
        ross & 7,259 & 70.8& amy & 1,699& 42.1 \\
        monica & 6,752& 64.2 & howard & 4,013 & 36.4 \\
        \bottomrule
    \end{tabular}
    }
    \caption{\small{Accuracy decomposed across characters. We also provide the number of training utterances for each character.}}
    \label{tab:char_accuracy}
\end{table}

\begin{figure}[t!]
    % \fontsize{14}{10}\selectfont
    \centering
    \includegraphics[width=0.50\textwidth]{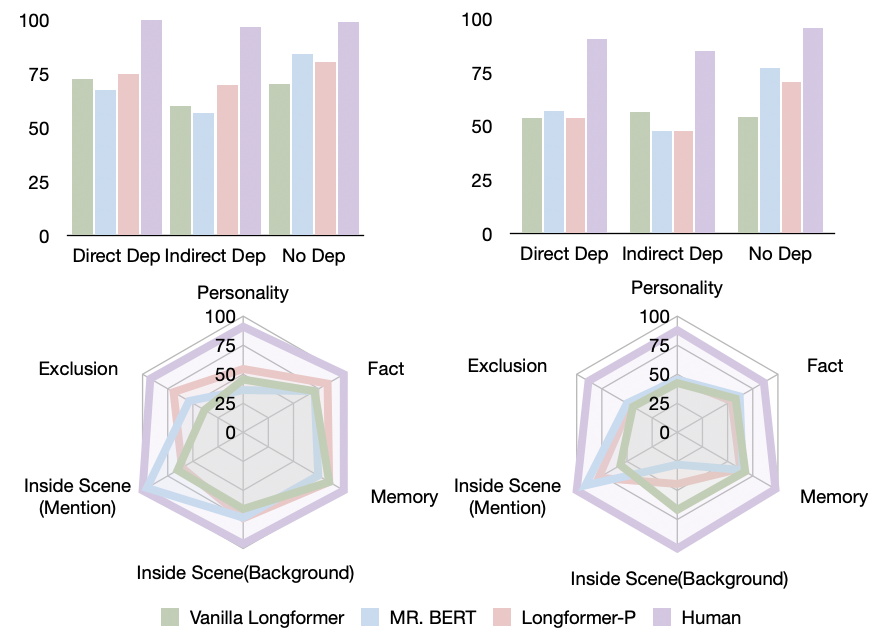}
    \vspace{-6mm}
    \caption{\small{Performance breakdown of Friends (the left column) and The Big Bang Theory (the right column). 
    % Note that human performs equally well across all the types.
    }}
    \label{fig:performance_breakdown}
    \vspace{-6mm}
\end{figure}

\medskip
\noindent\textbf{Impact of the Dependence on History }
The bar charts in Figure~\ref{fig:performance_breakdown} show the performance on different history dependence types.
The performance of cases that require history supports is in general harder for most of the models ($\sim$20\% lower compared to the cases without dependency of history).

The results suggest that a more thorough extraction of historical events associated with each character is needed to make the model an improvement. This notion aligns with the theories that past experience is an key aspect of building characters' ToM, showing that our \datasetname may serve as a good benchmark for the in-depth study of character comprehension from stories.

Another finding is that the cases requiring indirect history dependence (usually \emph{Personality} and \emph{Facts}) are even more challenging. 
The human mind develops detailed profiles of characters when reading stories.
The neural models represent each character as a single vector (\emph{i.e.}, the weight vector in the output layer).
This indicates a promising future direction of constructing structured persona representations (\emph{e.g.}, based on our schema of evidence) for more accurate character modeling.

\medskip
\noindent\textbf{Breakdown to Evidence Types } 
The wind-rose charts (bottom) in Figure~\ref{fig:performance_breakdown} provide performance breakdown onto our evidence categories. We omit the type of \emph{Linguistic style} because there are only two cases in \texttt{Friends} so the results are not stable.

As expected, the cases that can be resolved locally without character understanding (\emph{Inside-Mention}) are relatively easier.
All of \emph{Personality}, \emph{Fact} and \emph{Memory} cases have much lower performance as they correspond to heavy dependency on the modeling of history.

The type \emph{Exclusion} gives the worst overall performance on the two shows.
However, this does not indicate difficulty of character understanding -- According to the definition, these cases cannot be directly resolved with the scene inputs, but require the model to have specific strategy to exclude some incorrect answers first.

It is surprising that the \emph{Inside-Background} type poses difficulties to our models, because it looks to human annotators mostly like standard textual inference.
We have identified two possible reasons: 
(1) As discussed in the introduction, some cases require pragmatic understanding from the surface form to intention, only after which textual inference can be performed.
(2) The proportion of instances of this type is relatively smaller so the model may fail to recognize the required textual inference skills during training.

\medskip
\noindent\textbf{Effect of Scene Contexts }
Finally, the vanilla character classifier behaves differently  compared to the other models. Because it cannot make use of contexts within scenes, there is a performance gap on the  \emph{Inside-Mention} type (hence the drop on the \emph{No Dep} type).
However, it does not suffer from significant differential on the other types. The gap appears because Longformer-P and/or Mr. BERT perform considerably better on this type.

\medskip
\noindent\textbf{Challenges of History Retrieval}
Our experiments show that the history dependency presents serious challenges for existing models.
Finding the evidence from history scenes is a retrieval task (but without groundtruth).
We applied a state-of-the-art model to retrieve the history scenes and conducted an additional human study to evaluate the results.
Our study shows that on our identified cases with \emph{Direct Dependency}, the top-3 results (from in total 20 candidates) of a state-of-the-art semantic search model only give a recall of 35.5\%.
The result confirms that our task requires further advances on semantic retrieval.
The detailed task setting and our discussions can be found in Appendix~\ref{app:retrieval}.

\section{Conclusion}

In this paper, we present the first task and dataset for evaluating machine reading comprehension models for understanding characters in narratives. Based on linguistic, educational, and psychological theories, we proposed a new schema and conducted two human studies to analyze the types of evidence and reasoning required in understanding characters. We further design a new model architecture and conduct comprehensive experiments to serving as a testbed for future studies.

% We conduct a comprehensive analysis on the Book QA task, taking the representative NarrativeQA dataset as an example.
% Firstly, we design the Book QA techniques by borrowing the wisdom from the cutting-edge open-domain QA research and demonstrate through extensive experiments that (1) evidence retrieval in Book QA is difficult even with the state-of-the-art pre-trained LMs, due to the factors of rich writing style, recurrent book plots and characters, and the requirement of high-level story understanding; (2) our proposed approaches that adapt pre-trained LMs to books, especially the prereading technique for the reader training, are consistently helpful. 

% Secondly, we perform a human study and find that (1) a majority of questions in Book QA requires understanding and differentiating events and their relations; (2) the existing pre-trained LMs are deficient in extracting the inter- and intra-structures of the events in the Book QA. Such facts lead us towards the event understanding task for future improvement over the Book QA task.

%%%%%%%%%%%%%%%%%%%%%%%%%%%%%%%%%%%%%%
%%%%%%%%%   Acknowledgment   %%%%%%%%%
%%%%%%%%%%%%%%%%%%%%%%%%%%%%%%%%%%%%%%

\bibliographystyle{acl_natbib}
\bibliography{custom}

\clearpage

\appendix
\section{A Detailed Survey of Related Work}
\label{app:survey}

In this section, we first give an in-depth analysis on the difference between narrative and synopsis, from both the empirical challenges in NLP studies and the linguistic theory from~\cite{morrow1985prominent}.
Then we provide detailed discussion on how we summarize related work in Table~\ref{tab:existing_ds2}.

\begin{table*}[t]
\small
\centering
\resizebox{\textwidth}{!}{%
\begin{tabular}{lcccccccccc}
\toprule
\multirow{2}{*}{\textbf{Dataset}}& \multirow{2}{*}{\textbf{Task Format}} & \multicolumn{2}{c}{\textbf{Narrative Type}} & \multicolumn{3}{c}{\textbf{Assessed Narrative Comprehension Skills}} & \multicolumn{3}{c}{\textbf{Assessed Commonsense Knowledge}}\\
& & \textbf{Source} & \textbf{Length} & \textbf{Plot Structures} & \textbf{Character Facts} & \textbf{Character ToMs} & \textbf{Concepts} & \textbf{Events/States} & \textbf{Story Flows}\\
\midrule
\textbf{MCTest} &  Multi-choice QA & {\begin{tabular}[c]{@{}c@{}}Short fiction\\ \scriptsize{(Children stories)}\end{tabular}} & $\sim$20$^\ast$ & \checkmark &  &  & \checkmark & \checkmark & \checkmark \\
\textbf{BookTest} & Cloze test & {\begin{tabular}[c]{@{}c@{}}Literature\\ \scriptsize{(Excerpt)}\end{tabular}} & - & \checkmark & &  \\
\textbf{\cite{ma2018challenging}} & Cloze test &{\begin{tabular}[c]{@{}c@{}}TV show transcripts\\ \scriptsize{(Scenes)}\end{tabular}} & $\sim$20 & \checkmark & \\
% \textbf{NarrativeQA} & Generative QA & {\begin{tabular}[c]{@{}c@{}}Movie Scripts, Literature\\ \scriptsize{(Full stories)}\end{tabular}} & $\sim$$1$k(movie),$\sim$$11$k(book)\ast & \checkmark & \checkmark & & & \checkmark  \\
\textbf{NarrativeQA} & Generative QA & {\begin{tabular}[c]{@{}c@{}}Movie Scripts, Literature\\ \scriptsize{(Full stories)}\end{tabular}} & $\sim$11K$^\ast$ & \checkmark & \checkmark & & & \checkmark  \\
\textbf{FriendsQA} & Extractive QA &{\begin{tabular}[c]{@{}c@{}}TV show transcripts\\ \scriptsize{(Scenes)}\end{tabular}} & $\sim$20$^\ast$  & \checkmark & \checkmark\\

\textbf{TellMeWhy} & Multi-choice QA &{\begin{tabular}[c]{@{}c@{}}Short fiction\\ \scriptsize{(ROCStories)}\end{tabular}} &  5 &  &   &  & & \checkmark  \\

% \textbf{RACE} & Multi-choice & {\begin{tabular}[c]{@{}c@{}}(Partially) Literature\\ \scriptsize{(Short story or excerpt)}\end{tabular}} & Natural & Free-form & Expert & \\
\textbf{NovelChapters/BookSum} &Summarization &{\begin{tabular}[c]{@{}c@{}}Literature\\ \scriptsize{(Chapters or Full stories)}\end{tabular}} &$\sim$4K&\checkmark&&&&&\checkmark\\
\textbf{SummScreen} &Summarization &{\begin{tabular}[c]{@{}c@{}}TV show transcripts\\ \scriptsize{(Scenes)}\end{tabular}} &$\sim$330&\checkmark&&&&&\checkmark\\
{\begin{tabular}[c]{@{}c@{}}\textbf{\cite{chen2016character} /} \\ \textbf{\cite{chen2017robust}}\end{tabular}} & Coref Resolution &{\begin{tabular}[c]{@{}c@{}}TV show transcripts\\ \scriptsize{(Episodes or scenes)}\end{tabular}} & $\sim$20/260$^{\dagger}$ & \checkmark & \checkmark & & & \checkmark &\checkmark  \\
\textbf{\cite{flekova2015personality}} & Classification &{\begin{tabular}[c]{@{}c@{}}Literature\\ \scriptsize{(Full stories)}\end{tabular}} & $\sim$22K  & &\checkmark & \\
% \textbf{\cite{flekova2015personality}} & Classification &Full stories&&&\checkmark&&&&&\checkmark\\
\midrule
\textbf{\datasetnamens} & Multi-choice & {\begin{tabular}[c]{@{}c@{}}TV show transcripts\\ \scriptsize{(Full stories)}\end{tabular}} &  $\sim$50K & \checkmark \scriptsize{(indirect)} & \checkmark & \checkmark & \checkmark & \checkmark & \checkmark \\
\bottomrule
\end{tabular}
}
\caption{\small{Properties of existing narrative comprehension datasets compared to \datasetname. 
We organize the datasets according to the following dimensions related to narrative understanding:
% \textbf{focusing more on the coverage of reading skill types}
\textbf{Source} of the texts for reading comprehension;
\textbf{Length} of the texts from the source that makes the task solvable, we report the numbers of sentences or utterances for books and scripts respectively;
whether the task assesses the ability of understanding \textbf{plot structures} in the stories;
whether the task assesses the ability of understanding basic \textbf{character facts} like personality, profession, etc;
whether the task assesses the ability of building \textbf{character theory-of-mind (ToM)}; whether the task assesses the commonsense knowledge of \textbf{concepts}, \textbf{events} and \textbf{states}; and whether the task assesses the additional commonsense about the \textbf{narrative development}, including the knowledge about the coherence among non-verbal narratives and dialogues, and how they form the story/plot flow.
* Numbers are not reported in the original paper so we calculated them from the dataset.
$\dagger$\cite{chen2017robust} proposes two settings with single scene and the whole episode as inputs respectively. Different from ours, their include of episode is not to support the in-scene prediction with necessary history, but mostly increase the difficulty level of the co-ref task.}}
\label{tab:existing_ds2}
\end{table*}

\subsection{Background: Narrative versus Synopsis}
\label{app:synopsis_clarify}
As our work focuses on narrative comprehension, following  setups like~\cite{kovcisky2018narrativeqa,kryscinski2021booksum,chen2021summscreen}, it is necessary to distinguish between comprehension of  original narrative stories versus comprehension of their synopses (the human-written plot summaries), \emph{e.g.}, from the story's Wikipedia page.

Narrative stories are told by creating scenes, with the goal of helping readers  experience events as they occur in the plot, and empathize with the story characters in relation to their own experiences.
To engage  readers, story writers often use complex narrative clues (\emph{e.g.}, character activities, event development, scenery changes); variable narrative sequences (\emph{e.g.}, narrative, flashback, interpolation); and various linguistic expressions (\emph{e.g.}, argument, lyricism, narrative, description, illustration).
By comparison, a synopsis is a descriptive summary of the main idea of a story in plain language. It contains only the main characters, time, place, important plot points, and resolution, rather than allowing the story to unfold through the actions of the characters. The goal of the synopsis is to inform readers what happened with little or no original material from original story.

Therefore, comprehension of narrative stories requires more sophisticated skills to understand the complex chain of clues and expressions, in order to finally build a complete narrative representation from a sequence of individual scene comprehensions along with a developed  understanding of characters' mental models~\cite{morrow1985prominent}. In this light, a synopsis represents "processed results" from the application of these comprehension skills by a (experienced) human reader.

\subsection{Background: Dialogue in Stories vs. Real-Life Conversation}
Fictional dialogue canonically serves a purpose in a narrative. Either i contributes to the develop of a character or advances the plot~\cite{mchale2004free}. Nash proposed three categories of fictional dialogue: (1) confrontational dialogue which ``includes challenges, quarrels, disputes, interviews, and any kind of personal encounter in which the participants are in covert or overt opposition to each other''; (2) instructional dialogue, which ``conveys information about matters of science, technology, politics, world events, etc, some knowledge of which is essential to understanding the plot''; (3) collaborative dialogue that consists of  ``a series of exchanges which cumulatively present, for the reader's benefit, a picture of events, histories, personalities, and relationships''~\cite{nash1990language}. Authors have total control of the fictional dialogue, and ideally it functions according to the author's intention at every part of the story~\cite{de1996more}. However, real-life conversation is a joint action and is natural. There is no individual has whole control and the conversation goal of each party may be very different. Additionally, real-life conversations are temporally linear, such that communicators cannot revise earlier speech to fit a story. Moreover, real-life conversations tend to be highly implicit because spoken language derives much of its meaning from context~\cite{warren2006features}.

\subsection{Assessment of Narrative Comprehension}
We summarize the related tasks people use for assessment of general narrative comprehension skills.

\paragraph{Cloze Test}
Cloze tests take a snippet of the original text with some pieces (usually entities) masked as blanks, with the goal of filling these blanks from a list of candidates.
Cloze tests can be automatically constructed, resulting in an advantage in creation of large scale datasets.
Examples of cloze tests for narrative comprehension assessments are BookTest~\cite{bajgar2016embracing} and \cite{ma2018challenging}.
Both datasets are based on excerpts of books or scenes of TV shows. As the input consists only of short paragraphs, there is not sufficient information to infer complex character set via reading the stories. Therefore, these datasets address few questions related to the understanding of characters.\footnote{There may be a possible confusion of these tasks and ours, as cloze tests also include filling in anonymous character names in the blanks.
However, in these tasks, the required answers are also anonymized character IDs that appear in the inputs, and the IDs for the same character are random across different scenes. Therefore the character's information is not available for learning by design. In other words, their design of tasks \emph{deliberately prevent} the task of character understanding.}

Moreover, when built on short snippets, cloze tests are known to prone to mostly local inference but not much reasoning and commonsense knowledge, as  studies in the NLP community suggested~\cite{chen2016thorough}. On the other hand, although our task also has form similar to cloze, it requires information about the characters from previous scenes, which is not only about understanding the characters, but also requires global inference across features of the story (see~Figure \ref{fig:bookqa_example}).

\paragraph{Question Answering}
The most popular form of narrative comprehension evaluation is through question answering, starting from the early work of MCTest~\cite{richardson2013mctest}, to the more recent crowd-sourced tasks like NarrativeQA~\cite{kovcisky2018narrativeqa}, FriendsQA~\cite{yang2019friendsqa}, TellMeWhy~\cite{lal2021tellmewhy} and FairytaleQA~\cite{xu2022fantastic}.

Among them, the MCTest and TellMeWhy conduct multi-choice question answering on short stories. As above, te input consists of short paragraphs, so there is not sufficient information to infer complex character factes via reading the stories. Therefore, these datasets also cover few questions assessing the understanding of characters.
The TellMeWhy has a specific focus on \emph{why}-questions assessing the causal knowledge between states and events. The inputs are short stories from the ROCStories dataset~\cite{mostafazadeh2016corpus}.
MCTest covers a much wider set of reading skills, as it is based on complete stories, and generates questions with the goal of assessing children's reading comprehension over both story plots and commonsense.

NarrativeQA and FriendsQA conduct natural question answering tasks.
NarrativeQA aims to infer free-form answers to questions about a specific book or movie script.
According to the human study from~\cite{mou2021narrative}, the major part of the dataset is event-centric questions, which queries the explicit plots from the original books thus do not require a significant amount of commonsense reasoning. The study also reveals that NarrativeQA consists of a small portion of character-related questions.
These questions mainly query the simple facts of characters, such as age and profession. The more complexity character persona types, like personality, emotional/psychological status and history experience studied in our work, are not covered.
Similar to NarrativeQA, FriendsQA is a QA task over TV show scripts. The dataset consists of six types of questions: \emph{who, what, when, where, why}, and \emph{how}. The \emph{who} questions target on asking speaker names of utterance contents or participants of events, therefore are mainly assessing understanding of plot structures (\emph{i.e.}, participant arguments of events).

Both NarrativeQA and FriendsQA have human-written questions with a reference of the plot summary, which require evidence explicitly exists in the original story texts, and thus do not have much requirement of reasoning.
The FriendsQA questions are based on scene summaries, and thus require mostly local evidence; the NarrativeQA questions are based on the book-level summary, and thus sometimes require the ability to bridge the gap between coarse-grained and fine-grained event descriptions (\emph{i.e.}, commonsense of sub-events).

\paragraph{Summarization} There is a recent trend to evaluate model's understanding of stories via
summarization, including NovelChapters~\cite{ladhak2020exploring}, BookSum~\cite{kryscinski2021booksum} and ScreenSum~\cite{chen2021summscreen}.
These works provide a good research opportunity to future story reading research, by showing that book-level or chapter-level summarization is challenging to existing machine reading models.
However, it is more difficulty to identify the specific required reading skills by these tasks, as 
there exist many factors beyond reading skills to generate a good summary, such as encoding and generating long narrative texts. 
Intuitively, story summarization is largely plot-related instead of character-related; and requires the knowledge to understand the story flow. 

\paragraph{Interactive Fiction Game Playing}
\textcolor{black}{Interactive fiction (IF) games~\cite{hausknecht2019interactive} have been proposed as a reinforcement learning task that requires understanding of narrative fiction stories as environment observations. Research work has successfully demonstrated that reading comprehension can provide helpful inductive biases for efficient policy learning~\cite{guo2020interactive}; while \citet{yao2021reading} also reveal the shortcomings of these games.
The debate calls for future investigations to understand the necessary narrative elements and the roles they play in the IF games.}

\subsection{Character-Centric Prediction over Narratives}
Our task can be seen as a character-centered understanding of the narrative, where the understanding of the character deepens the understanding of the story and makes the narrative engaging. There are limited studies on understanding characters' persona from reading stories. In this section we review some existing character-centric prediction tasks over narrative texts, and discuss the relations and differences.

\paragraph{Character Name Linking}
The task of coreference resolution for story characters~\cite{chen2016character,chen2017robust} is closely related to our \datasetnamens.
These coreference resolution focuses on identifying the characters mentioned in multiparty conversations from TV shows scripts.
The goal of these tasks is to resolve the coreference of pronouns and character-indicating nominals (\emph{e.g.}, \emph{you} and \emph{Mom}) \textbf{in dialogues} of the character names that appear in the local context. It also covers linking a named entity (\emph{e.g.}, \emph{Ross}) to the character, which is more on name matching instead of character understanding.

The task form of coreference resolution mainly requires the understanding of discourse relations. It does not assess the modeling of character theory-of-mind, especially the character's memories, as there are no predictions of character behaviors involved.
The major character persona type it assesses is character facts, since the resolution of nominals requires the understanding of the target characters' occupations and relationships.

The lack of ToM modeling and complex reasoning of the coreference resolution task also makes it relatively easier -- on \texttt{Friends} and \texttt{The Big Bang Theory}, a CNN model gives a $>$90\% average accuracy. By comparison, our task, although solvable by humans with a $\sim$95\% accuracy, is challenging to neural models as the best BERT-based model gives a $\sim$65\% average accuracy on the same two shows with even smaller candidate sets.

\paragraph{Personality Prediction}
Our work is also related to the prediction of fiction characters' personality types by reading the stories~\cite{flekova2015personality}. 
Specifically, the tasks require to predict a fiction character's MBTI personality types~\cite{myers1988myers} rooted in Jung's theory, based on the character's verbal and non-verbal narratives in the original stories.
Compared to the aforementioned character-centric prediction tasks, these studies require to read and comprehend the original long stories, but the prediction task are relatively simpler since they only focus on personality which is a single perspective of persona.

\subsection{Character-Centric Prediction over Non-Narratives}

\paragraph{Character name linking between story synopses}
Recently \citet{brahman2021let} propose the LiSCU, which is a novel textual entailment task linking an anonymous summative descriptions of story character to the name appearing in the story's plot summary.
Similarly to \cite{chen2016character}, the task assess the resolution of names and events instead of the ToM modeling. This is because the task does not involve much explicit behavior predictions, since the task form is entailment between two given statements rather than predicting the possibility of new contents.
The usage of synopses over original stories reduces the challenges in narrative understanding; and further prevents the character comprehension from stories, as pointed out by \cite{kovcisky2018narrativeqa}, the summaries themselves are humans' comprehension results of the stories.

\paragraph{Personalized Dialogue Generation}
Our work is also related to personalized dialogue generation, for which datasets~\cite{mairesse2007personage,walker2012annotated,zhang2018personalizing,li2020aloha} and models~\cite{li2016persona,mazare2018training,qian2018assigning,zheng2020pre} are proposed for generating dialogues for speakers with persona features.
These benchmarks usually cover a single aspect of the multi-dimensional persona~\cite{moore2017five}. For example, \texttt{PERSONA-CHAT}~\cite{zhang2018personalizing} focuses on personal facts such as ``\emph{I’m a writer}'' and ``\emph{I live in Springfield}'';
other works mainly focus on learning language styles from speakers' personality types, such as the Big Five traits of the extraversion personality in \texttt{PERSONAGE}~\cite{mairesse2007personage}, and the personality types derived from TV tropes (e.g. \textit{jealous girlfriend}, \textit{book doom}, \textit{anti hero}) in \texttt{ALOHA}~\cite{li2020aloha}. 

\texttt{LIGHT}~\cite{urbanek2019learning} is a crowd-sourced dataset for text game adventure research. It includes natural language descriptions of fantasy locations, objects and their affordances, characters and their personalities, dialogue and actions of the characters. The biggest difference between ours and LIGHT is that LIGHT is based on the local environment of the conversation, rather than on a story. Examples from the LIGHT dataset are independent conversations and the context in which they occur. Crowd workers created the dialogues of characters by a given setting and a persona. The persona is modeled by the Persona-Chat dataset which is defined as a set of three to five profile sentences describing their personal facts such as ``\emph{I am a part of a group of travelers}'' and ``\emph{I go from town to town selling food to the locals}''.  

To the best of our knowledge, none of the existing studies cover a comprehensive multi-dimensional persona similar to our work, especially with respect to how a character's past experience builds her ToM.

\paragraph{Authorship Attribution}
Finally, authorship attribution has parallel ideas to our task, insofar as it aims at guessing  author identities from the texts they wrote~\cite{ni2019justifying,andrews2019learning,bevendorff2020overview} and thus requires a certain degree of author profiling.
These tasks differ from ours because the reviews, tweets or fandoms under the same authors do not usually form consecutive plotlines. Therefore, the tasks mainly require to understand the authors' writing styles rather than building mental models from the their past experiences. From this perspective, this direction is in fact more closely related to stylistic analysis in narrative understanding~\cite{vishnubhotla2019fictional}, rather than character understanding.

% \clearpage

\section{Supplementary for the Dataset Analysis}
\label{app:annotation}
\subsection{Summary of the Annotation Schema}
We include a summary of our annotation schema in Figure~\ref{fig:evidence_category}.

\begin{figure*}[ht!]
    % \small
    \fontsize{8}{10.5}\selectfont
    \centering
    \begin{tabular}{lcp{11.8cm}}
        \toprule
        \multicolumn{1}{c}{\textbf{Evidence Type}} && \textbf{Description}\\
        \midrule
        
        \multirow{3}{*}{Linguistic style}
        && {Linguistic style refers to a character's individualized speech pattern. It consists of a selection of linguistic features such as vocabulary, syntactic patterns, rhythm, and tone. It also includes the use of elements such as direct or indirect, metaphor and irony.} \\
        \midrule
        
        \multirow{2}{*}{Personality}       
        & & {Personality is a person's stable attitude toward objective facts and the habitual way of behavior that is compatible with it. We adopt a wider definition of personal traits as in \cite{li2020aloha}.}\\
        \midrule
        
        \multirow{5}{*}{Fact}  &   \multirow{1}{*}{Attributes}
        & {Fact of a character's attributes in the TV series setting, such as race, profession, education level etc.} \\
        \cmidrule{2-3}
        &   \multirow{2}{*}{Relations}
        & {A character's relationship with others that truly exist in the TV series setting, including both social relations and drama role relations.} \\
        \cmidrule{2-3}
        &\multirow{1}{*}{Status}
        & {Facts of a character's temporal emotional or psychological status in the time period when the scene happens.}  \\
        \midrule
        \multirow{3}{*}{Memory}  &        
        & {The episodic memory about history events a character has in the previous show scenes. This also includes a rare case of a knowledge fact (i.e. the semantic memory) a character acquires from history scenes, which cannot be inferred from the facts of the character.}\\
        \midrule
        
        \multirow{3}{*}{Inside-scene}        & \multirow{2}{*}{Background} 
        & {The character's identity can be inferred from the background introduction of scene, or from the description of the other characters' words.}  \\ 
        \cmidrule{2-3}
        & \multirow{1}{*}{Mention} 
        & {The character's name or alias is called by the other people.} \\ 
        \midrule
        
        \multirow{2}{*}{Exclusion} 
        & &{The character's identity can be determined from the presence of characters in the scene and the other resolved characters.}\\ 
        
        \bottomrule
    \end{tabular}
    % \vspace*{-2mm}
    \caption{\small{The definitions of evidence types.}}
    \label{fig:evidence_category}
\end{figure*}
% \end{landscape}

\subsection{Examples of Each Evidence Types}
\label{app:evidence_type}

\paragraph{Linguistic Style}

    {\small
    \centering
    \begin{tabular}{p{7.3cm}}
      \rowcolor{Gray} {\textbf{Background:}} (from \texttt{TBBT}) \textit{[Amy's car]}\\
      \rowcolor{Gray} {\textbf{Candidates:}}  \textit{\{Leonard, Penny, Sheldon, Amy\}}\\
      \rowcolor{Gray} 
      \textbf{\textcolor{coutput}{P0}:} \textit{Whatever. You can't even go on a date without checking your relationship agreement.}\\
      \rowcolor{Gray} \textbf{\textcolor{cinput}{P1}:} \textit{If you've got a problem basing a relationship on a contract, I'd like to tell you about 13 plucky colonies that entered a relationship agreement called the U.S. Constitution. And it may not be cool to say so, but I think that love affair is still pretty hot today.}\\
      \rowcolor{Gray} \textbf{Answer: \textcolor{cinput}{P1}} $\rightarrow$ \emph{Leonard}\\ 
      \rowcolor{Gray} \textbf{Rationale:} (Shelton's language is characterized by the use of long, difficult sentences and references to historical stories.)
      \end{tabular}\\}
      
\paragraph{Personality} 
    {\small
    \centering
    \begin{tabular}{p{7.3cm}}
      \rowcolor{Gray} {\textbf{Background:}} (from \texttt{TBBT}) \textit{[The cafeteria]}\\
      \rowcolor{Gray} {\textbf{Candidates:}}  \textit{\{Leonard, Howard, Sheldon, Raj\}}\\
      \rowcolor{Gray} 
      \textbf{\textcolor{coutput}{P0}:} \textit{And you love the sound of your own voice.}\\
      \rowcolor{Gray} \textbf{\textcolor{cinput}{P1}:} \textit{Yeah, well, of course I do. Listen to it. It's like an earful of melted caramel.}\\
      \rowcolor{Gray} \textbf{Answer: \textcolor{cinput}{P1}} $\rightarrow$ \emph{Sheldon}\\ 
      \rowcolor{Gray} \textbf{Rationale:} (Sheldon is a self-centered person so he will praise his own voice.)
      \end{tabular}\\}
      
\paragraph{Memory}
    {\small
    \centering
    \begin{tabular}{p{7.3cm}}
      \rowcolor{Gray} {\textbf{Background:}} (from \texttt{TBBT}) \textit{[The stairwell]}\\
      \rowcolor{Gray} {\textbf{Candidates:}}  \textit{\{Leonard, Penny\}}\\
      \rowcolor{Gray} 
      \textbf{\textcolor{coutput}{P0}:} \textit{There's something I wanted to run past you.}\\
      \rowcolor{Gray} \textbf{\textcolor{cinput}{P1}:} \textit{What's up?}\\
      \rowcolor{Gray} \textbf{\textcolor{coutput}{P0}:} \textit{Mm, the guys and I were thinking about investing in Stuart's comic book store. Is that okay?} \\
      \rowcolor{Gray} \textbf{\textcolor{cinput}{P1}:} \textit{Why are you asking me?} \\
      \rowcolor{Gray} \textbf{Answer:} \textbf{\textcolor{coutput}{P0}} $\rightarrow$ \emph{Leonard}\\ 
      \rowcolor{Gray} \textbf{Rationale:} (In a previous scene, Leonard and his friends discussed about investing in Stuart's store, so he is the only one between the two who has this memory.)
      \end{tabular}\\}

\paragraph{Fact}
\begin{itemize}
\item{Attribute}

    {\small
    % \centering
    \begin{tabular}{p{7cm}}
      \rowcolor{Gray} {\textbf{Background:}} (from \texttt{TBBT}) \textit{[Amy's lab]}\\
      \rowcolor{Gray} {\textbf{Candidates:}}  \textit{\{Amy, Penny\}}\\
      \rowcolor{Gray} 
      \textbf{\textcolor{cinput}{P0}:} \textit{Hey. Ready to go to lunch?}\\
      \rowcolor{Gray} 
      \textbf{\textcolor{coutput}{P1}:} \textit{Just give me a minute. I'm stimulating the pleasure cells of this starfish. I just need to turn it off.}\\
      \rowcolor{Gray} 
      \rowcolor{Gray} \textbf{Answer:}\textbf{ \textcolor{coutput}{P1}} $\rightarrow$ \emph{Sheldon}\\ 
      \rowcolor{Gray} \textbf{Rationale:} (Sheldon is Amy's boyfriend. After identify P0 is Amy, based on the relationship between Amy and Sheldon, P1 can be identified as Sheldon.)
      \end{tabular}\\}
      
\item{Relationship}  

    {\small
    \centering
    \begin{tabular}{p{7cm}}
      \rowcolor{Gray} {\textbf{Background:}} (from \texttt{TBBT}) \textit{[Amy's lab]}\\
      \rowcolor{Gray} {\textbf{Candidates:}}  \textit{\{Amy, Penny, Sheldon\}}\\
      \rowcolor{Gray} \textbf{$\cdots$}\\
      \rowcolor{Gray} \textbf{\textcolor{coutput}{P0}:} \textit{Hey, boyfriend.}\\
      \rowcolor{Gray} \textbf{\textcolor{cinput}{P1}:} \textit{Can't talk. Spitball. Probably gonna die.}\\
      \rowcolor{Gray} \textbf{Answer:}\textbf{ \textcolor{cinput}{P1}} $\rightarrow$ \emph{Sheldon}\\ 
      \rowcolor{Gray} \textbf{Rationale:} (Sheldon is Amy's boyfriend. After identify P0 is Amy, based on the relationship between Amy and Sheldon, P1 can be identified as Sheldon.)
      \end{tabular}\\}
      
\item{Status}

    {\small
    % \centering
    \begin{tabular}{p{7cm}}
      \rowcolor{Gray} {\textbf{Background:}} (from \texttt{TBBT}) \textit{[The pub]}\\
      \rowcolor{Gray} 
      \textbf{\textcolor{coutput}{P0}:} \textit{So when do you guys plan on getting married?}\\
      \rowcolor{Gray} \textbf{\textcolor{cinput}{P1}:} \textit{Uh, we're not sure. But I want to wait long enough to prove to my mother I'm not pregnant.}\\
      \rowcolor{Gray} 
      \textbf{\textcolor{cgroundtruth}{P2}:} \textit{May I have one of your fries?}\\
      \rowcolor{Gray} \textbf{\textcolor{cinput}{P1}:} \textit{Of course. Can I have a bite of your burger?}\\
      \rowcolor{Gray} 
      \textbf{\textcolor{cgroundtruth}{P2}:} \textit{Absolutely not.}\\
      \rowcolor{Gray} \textbf{\textcolor{pink}{P3}:} \textit{Some perfect couple. He won't even share his food with her.}\\
      \rowcolor{Gray} \textbf{Answer:}\textbf{\textcolor{pink}{P3}} $\rightarrow$ \emph{Leonard}\\ 
      \rowcolor{Gray} \textbf{Rationale:} (The aforementioned failure to determine Leonard's marriage led him to ridicule couples in harmonious relationships.)
      \end{tabular}\\}

\end{itemize}

\paragraph{Inside-Scene} 
\begin{itemize}
\item{Background}
    {\small
    % \centering
    \begin{tabular}{p{7cm}}
      \rowcolor{Gray} {\textbf{Background:}} (from \texttt{TBBT}) \textit{[Penny's apartment]}\\
      \rowcolor{Gray} {\textbf{Candidates:}}  \textit{\{Amy, Penny\}}\\
      \rowcolor{Gray} 
      \textbf{Bernadette:} \textit{Nah, you got this. Let's go for a drink. \textbf{I'll call Amy}.}\\
      \rowcolor{Gray} \textbf{\textcolor{cinput}{P0}:} \textit{Okay, good. She seemed like she really wanted to go out tonight.}\\
      \rowcolor{Gray} \textbf{\textcolor{coutput}{P1}} (phone ringing, running down stairs from outside penny's door): \textit{Hey, girl.} \\
      \rowcolor{Gray} \textbf{Answer:}\textbf{ \textcolor{coutput}{P1}} $\rightarrow$ \emph{Amy}\\
      \rowcolor{Gray} \textbf{Rationale:} (Bernadette said she will call Amy and P1 is the person who answers the phone.)
      \end{tabular}\\}
      
\item{Mention}

    {\small
    % \centering
    \begin{tabular}{p{6.5cm}}
      \rowcolor{Gray} {\textbf{Background:}} (from \texttt{TBBT}) \textit{[The apartment]}\\
      \rowcolor{Gray} {\textbf{Candidates:}}  \textit{\{Raj, Leonard, Sheldon, Amy\}}\\
      \rowcolor{Gray} 
      \textbf{\textcolor{coutput}{P0}:} \textit{Mmm, I love how they put a waterfall at centre field. It really ties the whole stadium together.}\\
      \rowcolor{Gray} \textbf{\textcolor{cinput}{P1}:} \textit{This is fun, huh? We get to see our friend throw out the first pitch, have a hot dog, watch the game.}\\
      \rowcolor{Gray} \textbf{\textcolor{cgroundtruth}{P2}:} \textit{Whoa. Nobody said anything about watching the game.} \\
      \rowcolor{Gray} \textbf{\textcolor{brown}{P3}:} \textit{\textbf{Sheldon}, what did you expect?} \\
      \rowcolor{Gray} \textbf{Answer:} \textbf{\textcolor{cgroundtruth}{P2}} $\rightarrow$ \emph{Sheldon}\\
      \rowcolor{Gray} \textbf{Rationale:} (P3 mentioned the name of the person being questioned which is ``Sheldon'')
      \end{tabular}\\}

\end{itemize}

\paragraph{Exclusion}

   {\small
    \centering
     \begin{tabular}{p{6.5cm}}
      \rowcolor{Gray} {\textbf{Background:}} (from \texttt{Friends}) \textit{[Scene: Outside the Janitor's Closet, there are people having s*x and Mr. Geller is trying to give them some pamphlets.]}\\
      \rowcolor{Gray} {\textbf{Candidates:}}  \textit{\{Monica, Chandler\}}\\
      \rowcolor{Gray} 
      \textbf{Mr. Geller:} \textit{Kids, I spoke to a doctor and picked up this pamphlets on how to get pregnant. (He slides them under the door.}\\
      \rowcolor{Gray} \textbf{\textcolor{cinput}{P0}:} \textit{(walking by with Chandler.) Hey dad!}\\
      \rowcolor{Gray} \textbf{\textcolor{coutput}{P1}:} \textit{Hey.} \\
      \rowcolor{Gray} \textbf{Mr. Geller:} \textit{(pause) Sorry to bother you again, but could you pass my pamphlets back? (They do so.) Thank you.} \\
      \rowcolor{Gray} \textbf{Answer:} \textbf{\textcolor{coutput}{P1}} $\rightarrow$ \emph{Chandler}\\
      \rowcolor{Gray} \textbf{Rationale:} (Monica is Mr. Geller's daughter. P0 called Mr. Geller dad so she is Monica. There are only two candidate so the other one is Chandler)
      \end{tabular}\\}

% \clearpage
% \newpage
\section{Extended Study of Required Reasoning Types on our \datasetnamens}
\label{app:reasoning_type}
This section provides an in-depth analysis of the types of reasoning used to infer evidence when guessing characters.

\begin{figure*}[t!]
    % \small
    \fontsize{8}{10.5}\selectfont
    \centering
    \begin{tabular}{p{3cm}p{11.8cm}}
        \toprule
        \textbf{Reasoning Type} & \textbf{Description} \\
        \midrule
        \multirow{3}{*}{Default Conjunction}     
        &  {No single piece of evidence can solve the task, hence the conjunction among multiple pieces of evidence is required. This is the default reasoning type if there are multiple evidence types labeled but no other reasoning types are labeled.} \\
        \midrule
        
        \multirow{2}{*}{Multihop-Character}       
        & {Task needs to be solved with the guessing results of other characters, then using the target person relation to or memory about the guessed ones to make the answer, \emph{i.e.}, multihop with guessed characters as bridges.} \\
        \midrule
        \multirow{2}{*}{Multihop-Textual}          
        & {Task needs to be solved with the persona/fact/event not directly described in the scene but can be inferred from the context, \emph{i.e.}, multihop over persona/fact/event inferred from dialog and scene context.} \\
        \midrule
        \multirow{3}{3cm}{Commonsense attributes/relations of concepts/events}         
        & {Task requires additional commonsense knowledge of attributes of daily concepts or social events, or their relations like causal relations between events. Those refer to the specific types of commonsense covered in ConceptNet- or Atomic-style KBs.}  \\ 
        \bottomrule
    \end{tabular}
    % \vspace*{-2mm}
    \caption{\small{The definitions of reasoning types.}}
    \vspace*{-3mm}
    \label{fig:reasoning_category}
\end{figure*}
% \end{landscape}

\subsection{Our Annotation Schema of Reasoning Types}
We define the following reasoning types with examples provided. A summary of our annotation schema of reasoning types can be found in Figure~\ref{fig:reasoning_category}.
\paragraph{Multi-hop on Characters} 
Reasoning on the basis of other characters that have already been guessed. Using the already guessed character as a bridge, users can employ history event or the relationship between characters to make guesses about the target character.The difference between multi-hop character and exclusion is that after identifying the other characters, the exclusion technique relies only on the list of characters provided for guessing, however, multi-hop character reasoning requires additional evidence such as relationship to infer the target character.

{\small
    \centering
    \begin{tabular}{p{7.3cm}}
      \rowcolor{Gray} {\textbf{Background:}} (from \texttt{TBBT}) \textit{[Angels Stadium]}\\
      \rowcolor{Gray} {\textbf{Candidates:}}  \textit{\{Raj, Leonard, Sheldon, Amy\}}\\
      \rowcolor{Gray} 
      \textbf{\textcolor{coutput}{P5}:} \textit{Hey, I hear you're a dermatologist.}\\
      \rowcolor{Gray} \textbf{\textcolor{cinput}{Emily}:} \textit{Uh, yeah, I'm a resident at Huntington Hospital.}\\
      \rowcolor{Gray}
      ...\\
      \rowcolor{Gray} \textbf{\textcolor{coutput}{P5}:} \textit{I have some odd freckles on my buttocks. Can I make an appointment for you to look at them?} \\
      \rowcolor{Gray} \textbf{Emily:} \textit{Um, okay, I guess.} \\
      \rowcolor{Gray} \textbf{\textcolor{cgroundtruth}{P0}:} \textit{I'm with him three years, nothing. She's with him two minutes, and he's taking his pants off.} \\
      \rowcolor{Gray} \textbf{Answer:} \textbf{\textcolor{cgroundtruth}{P0} $\rightarrow$ \emph{Amy}}\\
      \rowcolor{Gray} \textbf{Rationale:} (Using P5 (Sheldon) as a bridge and the couple relationship between Amy and him, we can identify P0 is Amy.)
      \end{tabular}\\}

\paragraph{Multi-hop on Textual Evidence}
Some evidences are not directly presented in the scene but can be inferred from the descriptions of context and dialogues. Using the inferred evidences as bridges people can multihop  over  personality, or fact, or event inferred from the text to guess the characters.

{\small
    \centering
    \begin{tabular}{p{7.3cm}}
      \rowcolor{Gray} {\textbf{Background:}} (from \texttt{TBBT}) \textit{[The apartment ]}\\
      \rowcolor{Gray} {\textbf{Candidates:}}  \textit{\{Amy, Leonard, Raj, Howard', Penny, Sheldon\}}\\
      \rowcolor{Gray} 
      \textbf{Bernadette:} \textit{I like your suit.}\\
      \rowcolor{Gray} \textbf{\textcolor{cinput}{P0}:} \textit{Oh, thanks. Got a couple new outfits for work.}\\
      \rowcolor{Gray} \textbf{\textcolor{coutput}{P1}:} \textit{How does it feel knowing your fiancée's job is to go out and flirt with doctors, looking like that, while you sit here, you know, looking like this?} \\
      \rowcolor{Gray}
      ...\\
      \rowcolor{Gray} \textbf{Answer:} \textbf{\textcolor{cinput}{P0}} $\rightarrow$ \emph{Penny}\\
      \rowcolor{Gray} \textbf{Rationale:} (P0 has a new job can be inferred from the textual evidence ``Got a couple new outfits for work''. Plus we know that Penny has a new job, we can determine that P0 is Penny )
      \end{tabular}\\}

\paragraph{Commonsense of Concepts/Events}
Task requires additional commonsense knowledge of attributes of daily concepts or social events, or their relations including causal/effect relations between an event and a social state or social relation. We restrict this category to be the aforementioned commonsense knowledge types, to distinguish from other relatively under-studied commonsense knowledge, such as the commonsense of dialogue flow required to work with our inside-scene evidence defined in Figure~\ref{fig:evidence_category}.

    {\small
    \centering
    \begin{tabular}{p{7.3cm}}
      \rowcolor{Gray} {\textbf{Background:}} (from \texttt{TBBT}) \textit{[Capital Comics]}\\
      \rowcolor{Gray} {\textbf{Candidates:}}  \textit{\{Howard, Sheldon\}}\\
      \rowcolor{Gray} 
      ...\\
      \rowcolor{Gray} \textbf{\textcolor{cinput}{P0}:} \textit{I know that if I had a wife or a fiancée, I'd ask her first before I invested money in a comic book store.}\\
      \rowcolor{Gray} \textbf{\textcolor{coutput}{P1}:} \textit{He's right.} \\
      \rowcolor{Gray} \textbf{Answer:} \textbf{\textcolor{coutput}{P1}} $\rightarrow$ \emph{Howard}\\
      \rowcolor{Gray} \textbf{Rationale:} (A married or engaged person will answer ``He's right''.  Howard is married. )
      \end{tabular}\\}

\paragraph{Default Conjunction}
A single piece of evidence will not solve this task; a combination between multiple pieces of evidence is needed to identify the person.

\subsection{Analysis of the Human Annotation}

\paragraph{Correlation between the Human Annotated Schema Categories}
Figure~\ref{fig:annotation_category_flow} visualizes the flow between (a) evidence types and the dependency of history and (b) evidence types and the reasoning types. Most evidence types correlate with history dependency. Personality and history dependency are most closely related. Default conjunction is the reasoning type that accounts for the largest percentage.

\begin{table}[t]
\scriptsize
\centering
\small
\begin{tabular}{lccc} 
% \midrule \midrule
\toprule
Reasoning Type & Friends(\%) & TBBT(\%)         \\ 
\midrule 

        Default                & 16.56     & 28.48     \\
        Multihop(Character)    & 3.97      & 13.91     \\
        Multihop(Textual)      & 5.30      & 5.30      \\
        Commonsense            & 4.64      & 0.66      \\
        No Complex Reasoning   & 69.54     & 51.66     \\
\bottomrule
\end{tabular}

\caption{\small{Percentage of the required reasoning types in the two TV shows, \texttt{Friends} and \texttt{The Big Bang Theory}.}}
\label{tab:percentage_reasoning}
\end{table}

\begin{figure}[t!]
    % \fontsize{14}{10}\selectfont
    \centering
         \includegraphics[width=0.45\textwidth]{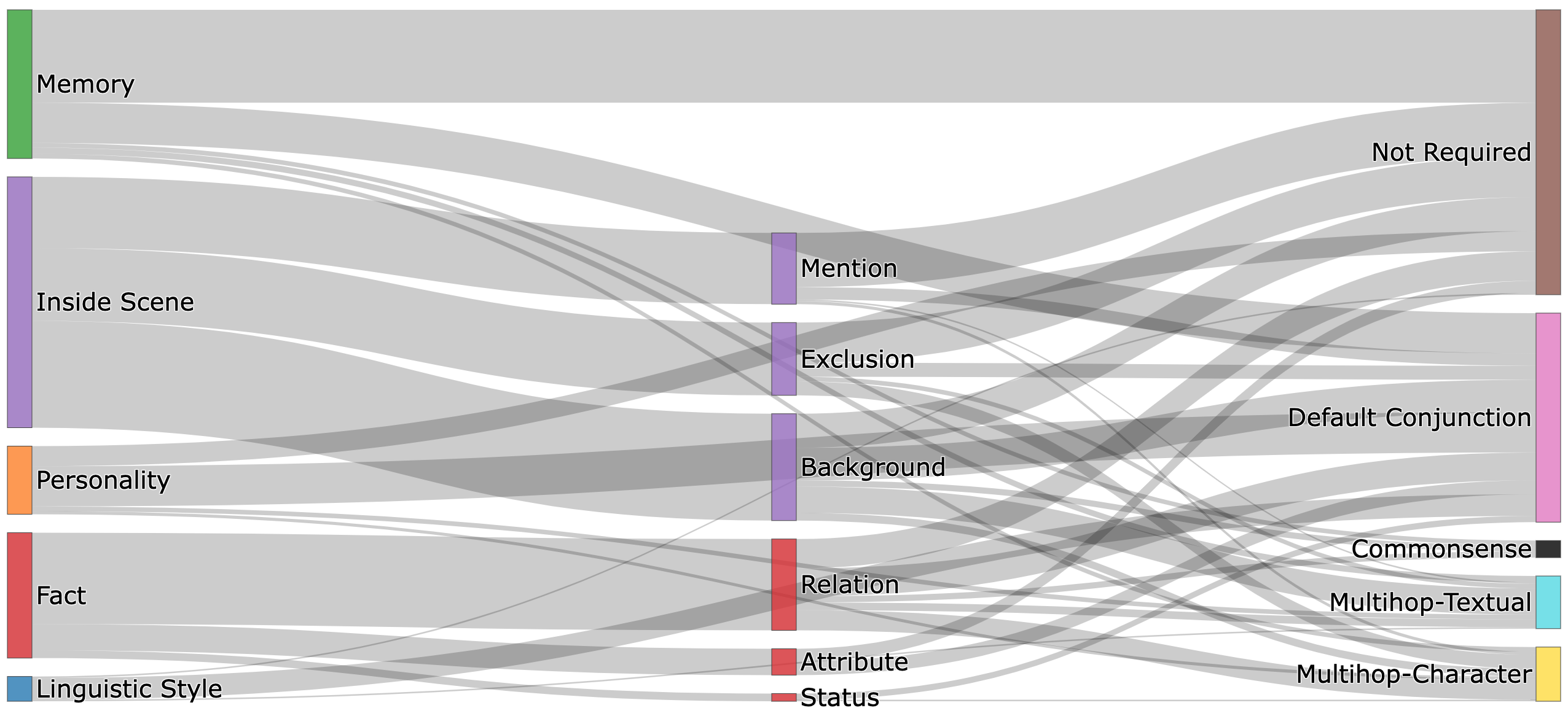}
    \caption{\small{Visualization of the flow from the required evidence types to their required reasoning types.}}
    \label{fig:annotation_category_flow_reasoning}
\end{figure}

\subsection{Experiments: Performance Decomposition on the Reasoning Types}
\begin{figure}[t!]
    % \fontsize{14}{10}\selectfont
    \centering
    \includegraphics[width=0.50\textwidth]{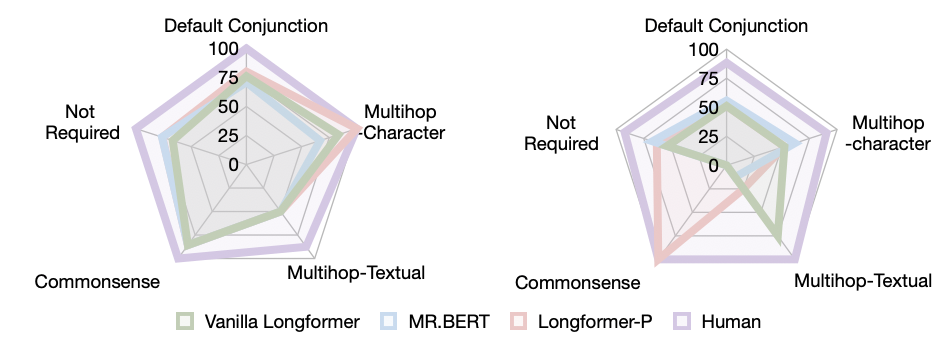}
    \vspace{-6mm}
    \caption{Performance breakdown according to our reasoning schema (left: Friends, right: The Big Bang Theory).}
    \label{fig:performance_breakdown_reasoning}
    % \vspace{-6mm}
\end{figure}

We further studied the impact of the required reasoning types on the performance (the right column in Figure~\ref{fig:performance_breakdown_reasoning}). In general there is a clear gap (on average $\sim$10\%) between cases that require complex reasoning and those that do not. The \emph{Multihop-Textual} type is most challenging, because it requires both deep understanding of what the texts implies and multihop reasoning. There is not a clear performance difference between \emph{Multihop-Character} and \emph{Default Conjunction}, though the former is conceptually harder. We surmise this is because both types are beyond the reasoning ability of the model so the predictions largely rely on fuzzy matching of evidence --  recall that we predict identities of main characters, so there can be a statistical bias of their context co-occurrence.
The results on the \emph{Commonsense} type fluctuate due to the relatively smaller ratio.

% \clearpage
\section{Interface for the Human Study}
\begin{figure}[th!]
    % \fontsize{14}{10}\selectfont
    \centering
    \begin{subfigure}[b]{0.5\textwidth}
         \centering
         \includegraphics[width=0.9\textwidth]{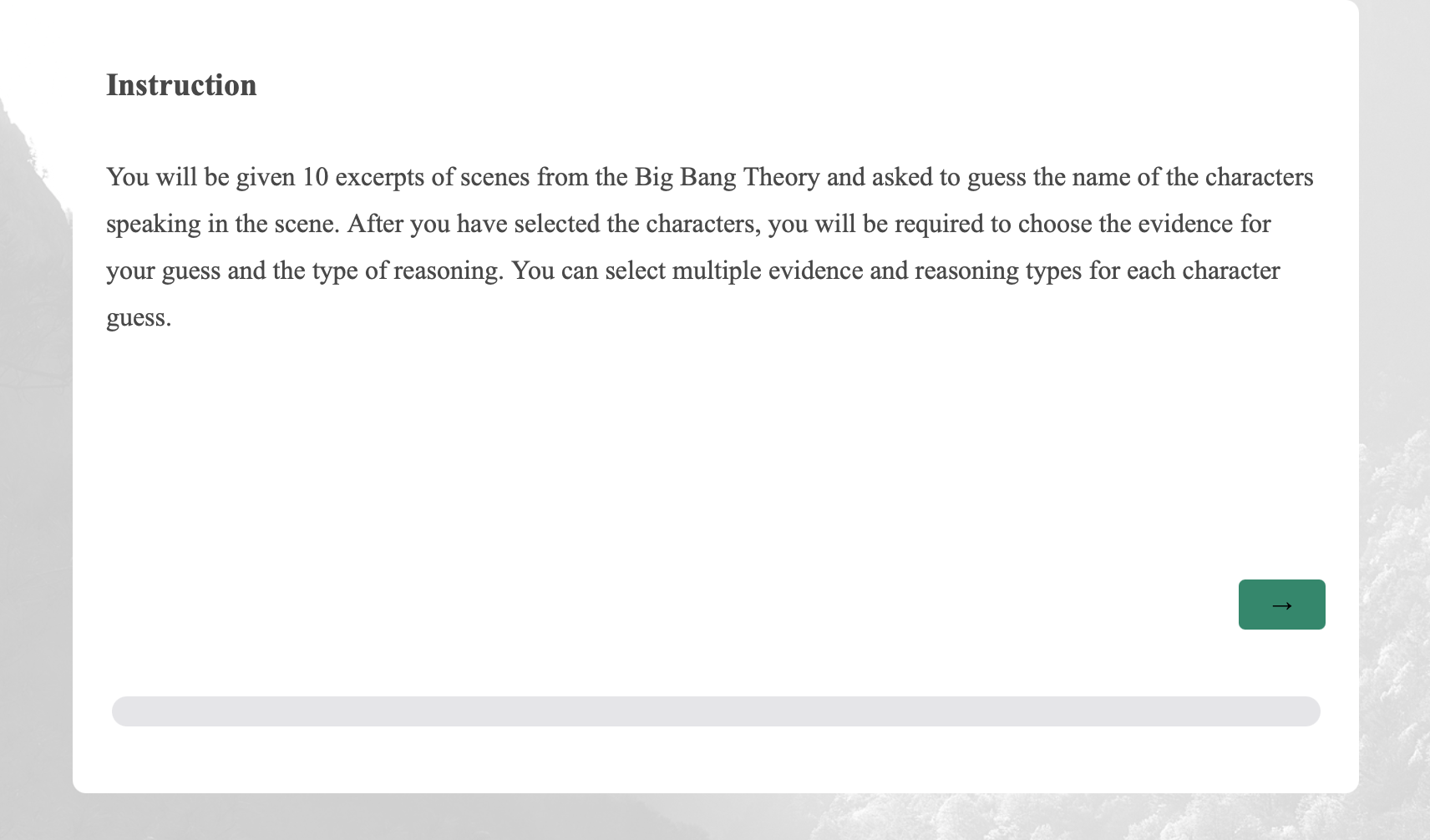}
         \caption{Introduction page of human study.}
        %  \label{fig:y equals x}
     \end{subfigure}
     \begin{subfigure}[b]{0.5\textwidth}
         \centering
         \includegraphics[width=0.9\textwidth]{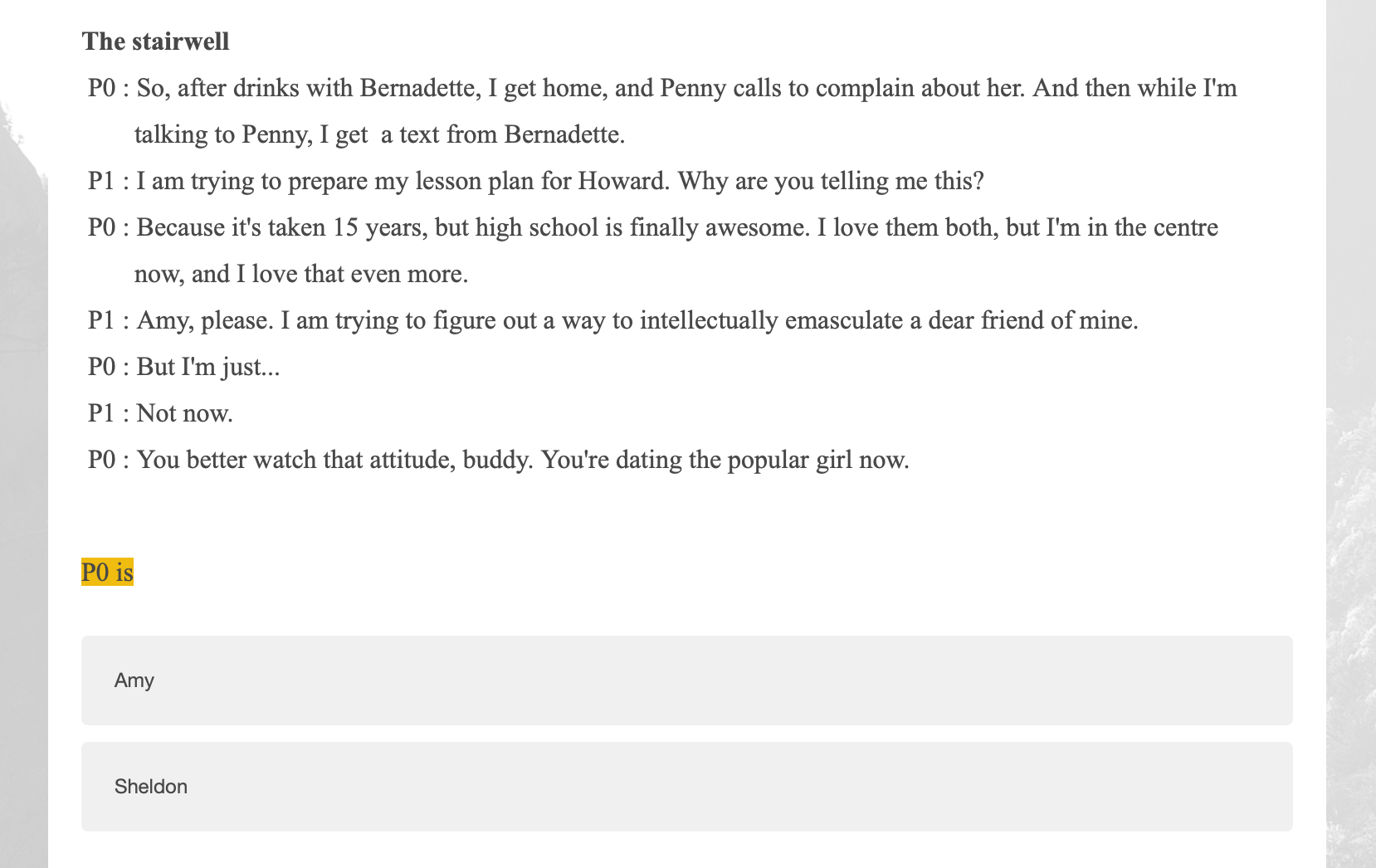}
         \caption{Task 1: character guessing task}
     \end{subfigure}
    % \vspace{-3mm}
        \centering
    \begin{subfigure}[b]{0.5\textwidth}
         \centering
         \includegraphics[width=0.9\textwidth]{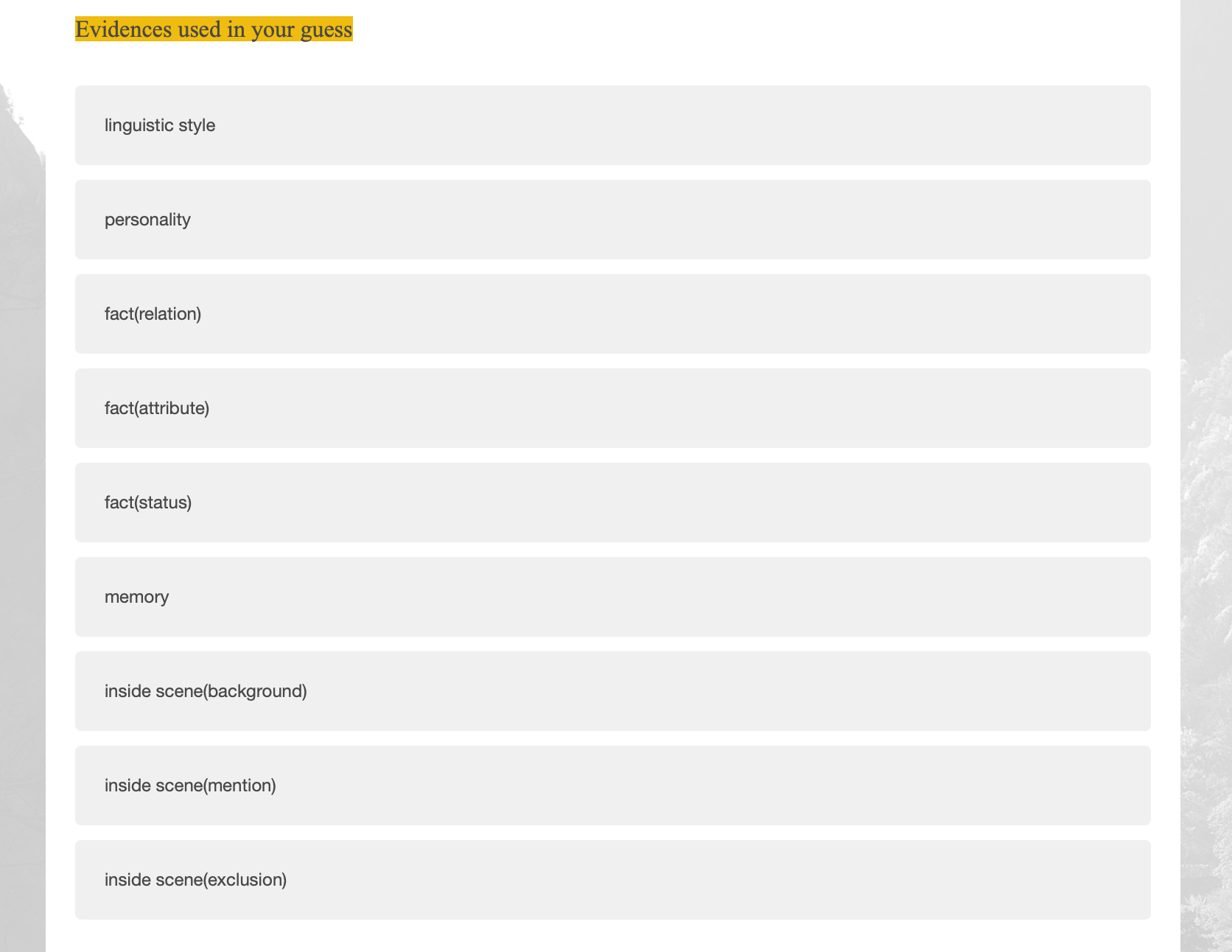}
         \caption{Task 2:identifying used evidence types.}
        %  \label{fig:y equals x}
     \end{subfigure}
         \centering
    \begin{subfigure}[b]{0.5\textwidth}
         \centering
         \includegraphics[width=0.9\textwidth]{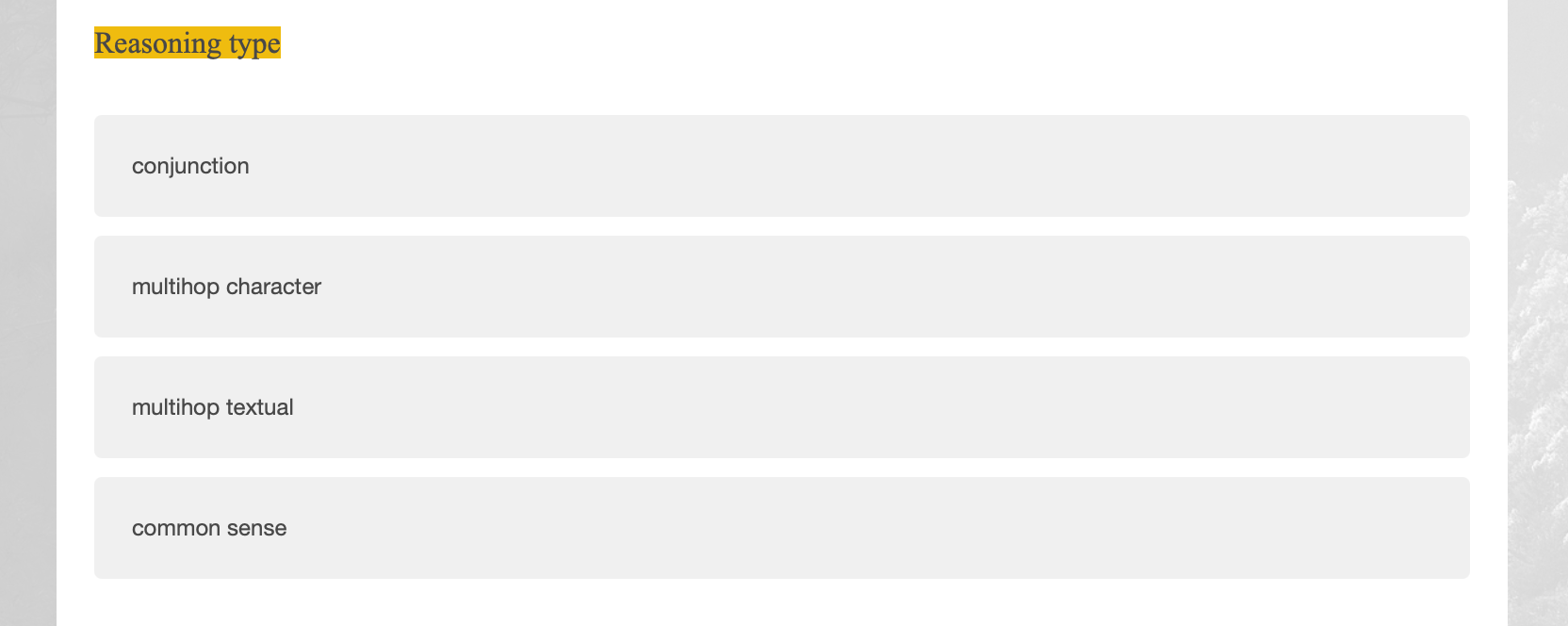}
         \caption{Task 3: identifying used reasoning types .}
     \end{subfigure}
    \caption{\small{interfaces of human studies.}}
    \label{fig: human_study_interface}
\end{figure}

Figure \ref{fig: human_study_interface} shows the interfaces of the human study.

% \clearpage
\section{Examples of Human Errors}
\label{app:human_error}
Table \ref{tab: unsolvable} provides an example of unsolvable cases and Table \ref{tab: mistake} provides an example of human mistakes. The human mislabeled characters are marked as red.

We further provide all the scene IDs on which our human tester makes incorrect predictions in Table~\ref{tab:human_error_cases}.

\begin{table}[!t]
    \small
    \centering
    \begin{tabular}{ll ll}
    \toprule
        \multicolumn{2}{c}{\bf \#Unsolvable} & \multicolumn{2}{c}{\bf \#Human Mistakes} \\ 
        \cmidrule(lr){1-2}
        \cmidrule(lr){3-4}
        TBBT & Friends & TBBT & Friends \\ 
        \midrule
        4882& 2500 &  4921  \\ 
        4895 &     & 4894  \\ 
        4907 &     & 4910 \\ 
        4908 & & \\
        \bottomrule
    \end{tabular}
    \caption{\small{Human Errors.}}
    \label{tab:human_error_cases}
\end{table}

{\begin{table*}
\small
\begin{center}
\begin{tabular}{||l||}
 \hline
   \textbf{Unsolvable Case} \\ [0.5ex] 
 \hline\hline
 08x02 4882\\
 \textbf{Background:} (from \texttt{TBBT}) \textit{[the Apartment]} \\ 
 \textbf{Candidates:} \textit{\{Howard, Sheldon, Raj, Amy, Leonard, Penny\}}\\
 P0 : I recently read that during World War Two, Joseph Stalin had a research program to create supersoldiers by having \\women impregnated by gorillas.\\
 P1 : What a sick use of science.\\
 P2 : Hey, as long as the baby's healthy.\\
 P3 : I wonder if Stalin considered any other animals.\\
 P4 : Hippos are the deadliest creature. A half-human, half-hippo
 soldier would be pretty badass.\\
 P1 : Yes, but when they're hungry-hungry, you can stop them with
 marbles.\\
 P0 : Yeah, the correct animal for interspecies supersolider is
 koala. You would wind up with an army so cute it\\ couldn't be
 attacked.\\
 P2 : But half-man, half-owl could fly...\\
 P0 : The answer is cuddly soldiers with big flat noses. Moving on.\\
 P1 : So, Penny, when's the new job start?\\
 P5 : Next Monday.\\
 Bernadette : Did you get a chance to look over the materials I gave
 you?\\
 P5 : Uh, not yet, but I will.\\
 Bernadette : Great. When?\\
 P5 : I said I'll get to it.\\
 P0 : I'm sensing awkwardness, am I right?\\
 P3 : Yes.\\
 P0 : Swish.\\
 Bernadette : I don't want to be pushy, but you've never done pharmaceutical sales before. 
 It seems like you could use this \\time to get a head start.\\
 P5 : Well, the first few weeks will be all training. They'll tell
 me everything I need to know.\\
 Bernadette : But imagine how impressed they'd be if you showed
 up already familiar with the material.\\
 P5 : Okay, so what, you want me to be like a teacher's pet?\\
 Bernadette : Couldn't hurt.\\
 P4 : Mm, I don't know. Who here has ever been hurt because they
 were the teacher's pet?\\
 P0 : It was like the rest of the class wanted Ms. McDonald to forget
 the quiz.\\
 \textbf{Answer: P0: Sheldon, \textcolor{red}{P1}: Howard, \textcolor{red}{P2}: Raj, P3: Amy, \textcolor{red}{P4}: Leonard, P5: Penny}\\[2ex] 
 \hline
\end{tabular}
\end{center}
\caption{Example of unsolvable case.}
\label{tab: unsolvable}
\end{table*}
}

{\begin{table*}
\small
\begin{center}
\begin{tabular}{||l||} 
 \hline
  \textbf{Mistake} \\ [0.5ex] 
 \hline\hline
 08x04 4921\\
 \textbf{Background:} (from \texttt{TBBT}) \textit{[Penny's partment]} \\ 
 \textbf{Candidates:} \textit{\{Raj, Penny\}}\\
 P0 : I'm so glad we could work this all out.\\
 P1 : Yeah, me, too.\\
 Emily : You know, we should have dinner one night with you and Leonard.\\
 P1 : Oh, we would love that.\\
 P0 : Great.\\
 background : (both chuckle)\\
 P1 : Okay, good night, guys.\\
 Emily : All right, night.\\
 P1 : Bye.\\
 Emily and Penny (simultaneously) : I hate her.\\
 \textbf{Answer: \textcolor{red}{P0}: Raj, \textcolor{red}{P1}: Penny}\\[2ex] 
 \hline
%  \label{tab: unsolvable}
\end{tabular}
\end{center}
\caption{Example of mistake.}
\label{tab: mistake}
\end{table*}
}

\section{Details of Human Study and Discussions on the Challenges of History Retrieval}
\label{app:retrieval}
Our experiments show that the history dependency challenges existing models.
Finding the evidence history scenes for such cases is essentially a retrieval task (but without groundtruth).
To see how it brings new challenges to existing semantic search, we applied a state-of-the-art model to retrieve the history scenes and conducted an additional human study to evaluate the results.

\noindent\textbf{Task }
We conduct the study on scenes in our human annotation sets that have the \emph{Memory} type labeled. 
With each scene as a query, we retrieve from a window of 20 previous scenes with a state-of-the-art model\footnote{We use the \texttt{all-mpnet-base-v2} model from \url{https://sbert.net/} that reports the top-1 performance on 14 sentence embedding tasks and 6 semantic search tasks.}
% The model is designed as general purpose model and trained on more than 1 billion training pairs.}.
The window size is decided so as to guarantee that at least one required memory appears in the window, according to our human annotation process.
The task of human study is to recognize whether the top-3 returned scenes contain at least one related history scene.

\noindent\textbf{Results }
The same annotators working on the study in Section~\ref{sec:analysis}  evaluated the retrieved scenes.
Results show that the recall of the top-3 results from this state-of-the-art model is low ({35.5\%}).
This difficulty in scene retrieval may arise from: (1) the queries are scenes with structures, which leads to different query formats from standard IR tasks; (2) many relevant scenes are dissimilar to the query scenes in the semantic space, but associated with the query in specific aspects or even analogous to the query scene; (3) some scenes require multi-hop retrieval, especially when combined with ToM modeling (reasoning about what others know).

All these challenges are non-trivial, and calls for further studies on semantic search to address.

\section{Model Checklist}
\label{app:checklist}

We implement our baselines based on HuggingFace Transformers.\footnote{https://github.com/huggingface/transformers} We use the pre-trained \texttt{allenai/longformer-base-4096} and \texttt{bert-base-uncased} models.
We train all the models with the Adam optimizer.

We train our model on a single V100 GPU.
It takes around 1 hour and 40 minutes to train a Longformer-based model.
It takes around 2 hour and 10 minutes to train a multi-row BERT model.
For all the models, we train in total 40 epochs. But the models usually converge in less than 20 epochs.

\paragraph{Hyperparameters}
We set the number of rows in MR. BERT to 12, to maximize the usage of GPU memory.
We set the maximum length of Longformer to 2000, which can handle the lengths of most of the input scenes. The window size is set to 256.
We set the learning rate to 2e-5.

We report our result with a single run. However, for each model we run twice; and we found the average development accuracy varies less than 0.5\%.

\end{document}